%% file: acl_latex_preprint.tex
\documentclass[11pt]{article}

\usepackage[preprint]{acl}

\usepackage{times}
\usepackage{latexsym}

\usepackage[T1]{fontenc}
\usepackage[utf8]{inputenc}

\usepackage{microtype}

\usepackage{inconsolata}

\usepackage{graphicx}

\usepackage[T1]{fontenc}    % use 8-bit T1 fonts
\usepackage{hyperref}       % hyperlinks
\usepackage{url}            % simple URL typesetting
\usepackage{booktabs}       % professional-quality tables
\usepackage{amsfonts}       % blackboard math symbols
\usepackage{nicefrac}       % compact symbols for 1/2, etc.
\usepackage{microtype}      % microtypography
\usepackage{xcolor}         % colors
\usepackage{longtable}
\usepackage{amsmath}
\usepackage{tabularx}
\usepackage{graphicx}
\usepackage{pifont}
\usepackage{multirow}
\usepackage{colortbl}
\usepackage[most]{tcolorbox}
\usepackage{listings}
\usepackage{wrapfig}
\usepackage{algorithm}
\usepackage{algpseudocode}
\usepackage{caption}
\tcbuselibrary{breakable} 
\usepackage{enumitem}
\usepackage{pifont}

\definecolor{openBg}{RGB}{254, 243, 232}    % 暖橙色
\definecolor{closedBg}{RGB}{234, 232, 248}  % 淡紫色
\definecolor{sharedBg}{RGB}{242, 242, 242}  % light gray
\definecolor{wrBg}{RGB}{232, 242, 255}      % light blue
\definecolor{lrBg}{RGB}{241, 235, 255}      % light purple
\definecolor{rpBg}{RGB}{235, 248, 238}      % light green
\definecolor{prBg}{RGB}{255, 239, 239}      % light red
\definecolor{closedFrame}{RGB}{120, 110, 170}   % 紫灰色边框
\definecolor{posGreen}{RGB}{0, 125, 60}
\definecolor{negRed}{RGB}{180, 35, 35}

\newcommand{\dpos}[1]{%
  \raisebox{-0.35ex}{\textcolor{posGreen}{\fontsize{5.5pt}{6pt}\selectfont +#1}}%
}
\newcommand{\dneg}[1]{%
  \raisebox{-0.35ex}{\textcolor{negRed}{\fontsize{5.5pt}{6pt}\selectfont -#1}}%
}
\newcommand{\xmark}{\ding{55}}

\lstdefinestyle{promptstyle}{
    basicstyle=\ttfamily\footnotesize,
    breaklines=true,
    breakatwhitespace=false,
    columns=fullflexible,
    keepspaces=true,
    showstringspaces=false,
    frame=none,
    xleftmargin=2mm,
    xrightmargin=2mm,
    aboveskip=0pt,
    belowskip=0pt
}

\usepackage[textsize=tiny]{todonotes}

\title{CDR-Bench: Evaluating Faithful Execution of Compositional, Order-Sensitive Data Refinement Recipes}

\author{
    \small Yuchen Huang$^{1,3}$\thanks{Work done during internship at Alibaba Group.} \quad
    Xiang Li$^{2,3}$\footnotemark[\value{footnote}] \quad
    Zhenqing Ling$^3$ \quad
    Sijia Li$^1$ \quad
    Qianli Shen$^3$ \\
    \small {%
    \setcounter{footnote}{1}%
    \textbf{Daoyuan Chen}$^{3}$\thanks{Corresponding Authors.} \quad
    \textbf{Yi R. (May) Fung}$^{1}$\footnotemark[\value{footnote}]} \quad
    \textbf{Yaliang Li}$^3$ \\[0.5em]
    \small $^1$HKUST \quad $^2$NUS \quad $^3$Tongyi Lab~\includegraphics[height=0.9em]{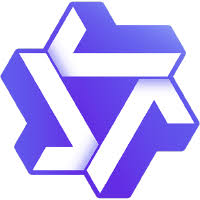}, Alibaba Group
}

\begin{document}
\maketitle
\begin{abstract}
% Data refinement pipelines routinely require chaining multiple operations over evolving text states, where both composition and execution order determine the final outcome. 
Data refinement involves executing multi-step recipes over evolving text states, where both composition and execution order of processing operators determine the outcome.
% Yet existing benchmarks either focus on isolated text editing or bundle refinement with code generation and tool invocation, leaving a core question unexamined: can an LLM faithfully execute a compositional, order-sensitive refinement recipe? 
% We introduce CDR-Bench, a benchmark of 3,319 tasks spanning four real-world data refinement domains and 29 operators, with deterministic reference outputs enabling exact evaluation without LLM judges. We evaluate three capability levels, namely atomic operator execution, order-agnostic recipe execution, and order-sensitive recipe execution, and propose three complementary metrics, Recipe Success (RS), Refinement Gain (RG), and Order-Consistent Success (OCS). 
While existing benchmarks either isolate text editing or entangle it with code and tool execution, it remains unclear whether LLMs can directly and faithfully execute these compositional, order-sensitive data refinement recipes. To fill this gap, we introduce CDR-Bench, a comprehensive benchmark featuring 3,462 high-quality tasks spanning four real-world data refinement domains and 29 distinct operators. 
Our benchmark evaluates models across atomic, order-agnostic, and order-sensitive settings, leveraging deterministic reference outputs to enable exact evaluation.
Experiments on 10+ state-of-the-art LLMs reveal consistent failure patterns: performance degrades sharply in compositional settings, and order-sensitive recipe success collapses.
% \textcolor{red}{One more finding here if possible.}
% Experimental results on 10+ state-of-the-art LLMs show . 
% Models degrade sharply from atomic to compositional settings. Under order-sensitive recipes, RS collapses while RG remains stable, indicating plausible-but-unfaithful outputs, and group-level OCS@3 never exceeds 4.90\% across all models. 
% State-Aware prompting and thinking mode both help but leave performance far below human-level, 
These findings underline that current LLMs lack the procedural faithfulness required for reliable compositional data refinement\footnote{Our code and data are released at \url{https://github.com/lukahhcm/data-juicer-hub/tree/CDR-Bench}.}.

\end{abstract}

\input{sections/intro}
\input{sections/method}

\input{sections/experiment}

\input{sections/background}

\input{sections/conclusion}

\section*{Limitations}
CDR-Bench has several limitations. First, it is grounded in the semantics
and coverage of the Data-Juicer operator library. To keep evaluation
deterministic, we exclude subjective refinement recipes and do not use LLM
judges as final evaluators, which limits coverage of broader real-world
refinement practices. Second, prompt verbalization partly relies on LLM
assistance. Although we average over multiple prompt styles and evaluate
several prompting strategies, this process may introduce two gaps: the
generated instructions may differ from how users naturally describe
refinement needs, and they may imperfectly verbalize the precise execution
semantics of the underlying deterministic recipes, despite validation.
Third, CDR-Bench focuses on direct text-level execution and does not cover
tool use, code generation, or interactive refinement in agentic settings.
Finally, multilingual and multimodal refinement remain outside the current
scope. We leave these directions for future work.

\section*{Ethics Statement}
CDR-Bench includes tasks involving privacy redaction and sensitive text cleanup. To mitigate potential risks, we rely on public, licensed, or synthetic data sources where appropriate. The released benchmark will avoid exposing raw private credentials, personal identifiers, or other unsafe content beyond what is already controlled in the source data. The goal of CDR-Bench is to improve the reliability of safe data processing and privacy-preserving refinement, rather than to facilitate misuse.

\section*{Reproducibility Statement}
To support reproducibility, we release benchmark instances, recipe metadata, prompt variants, evaluation code, and scripts for API and vLLM inference. Since gold references are derived from deterministic operator execution, future work can reproduce the evaluation without relying on subjective model-based judging.

%\section*{Acknowledgments}

\bibliography{custom}

\appendix

\input{sections/appendix}

\end{document}

%% file: sections/intro.tex
\section{Introduction}

\begin{figure}
    \centering
    \includegraphics[width=\linewidth]{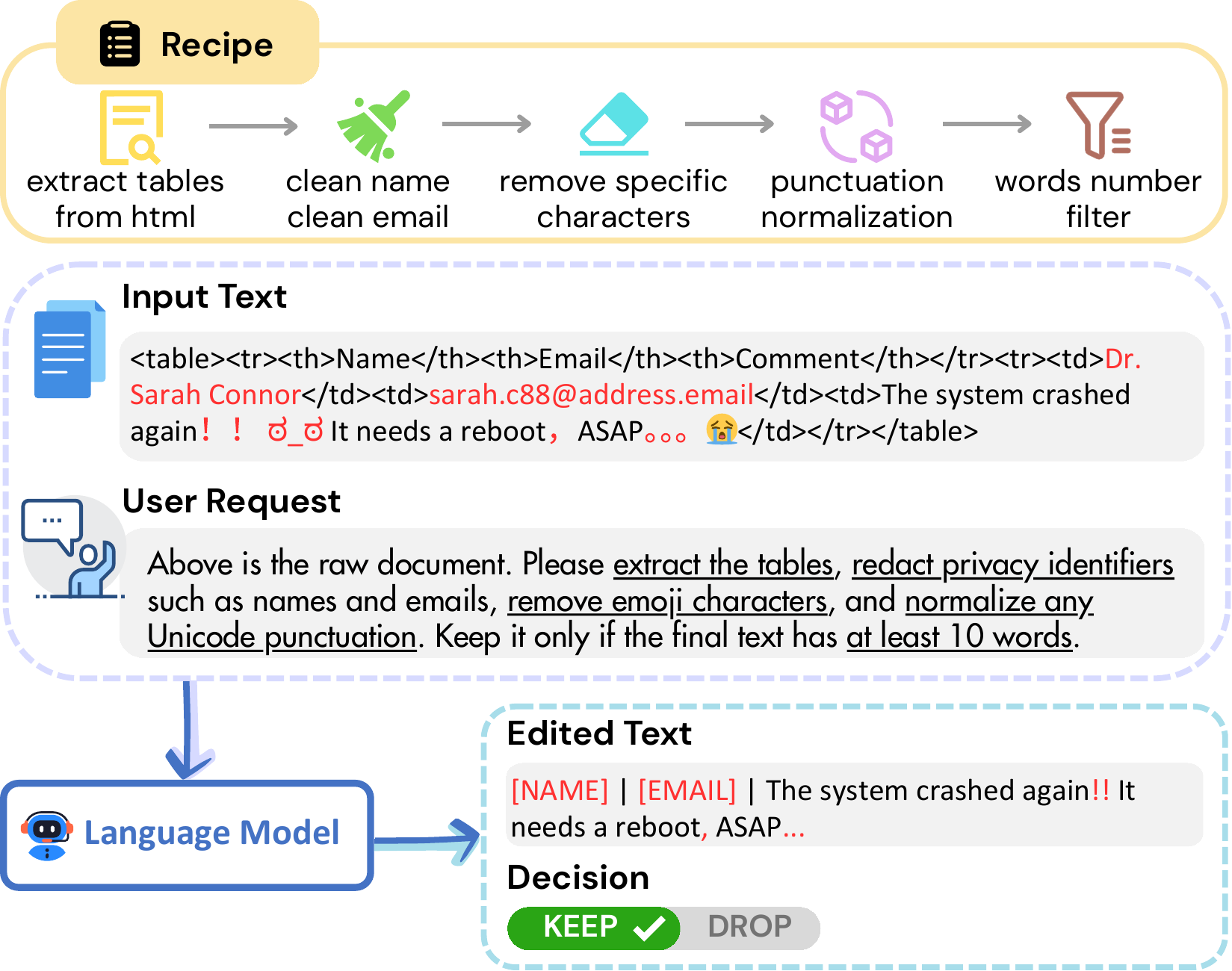}
    \caption{Illustration of a compositional data refinement process. The language model receives a raw input document alongside a user request specifying a multi-step recipe procedure, and directly outputs the processed target text together with a final execution judgment.}
    \label{fig:demo}
    \vspace{-1\baselineskip}
\end{figure}

Data refinement, the process of transforming noisy and heterogeneous raw text into clean, task-ready data, is a core component of modern LLM pipelines. It underpins applications such as pretraining corpus construction~\citep{djv1,qin2025scalinglawssyntheticdata}, reliable knowledge retrieval in RAG systems~\citep{liu2026tacklinginherentdifficultynoise,khan2024developingretrievalaugmentedgeneration}, and privacy-sensitive deployments~\citep{garza2025prvlquantifyingcapabilitiesrisks,pal2024empiricalimpactdatasanitization}. 
% Data refinement constitutes a cornerstone of the Large Language Model (LLM) lifecycle. It supports the construction of high-quality pretraining corpora~\citep{djv1,qin2025scalinglawssyntheticdata}, enables reliable knowledge retrieval in RAG systems~\citep{liu2026tacklinginherentdifficultynoise,khan2024developingretrievalaugmentedgeneration}, and protects privacy in sensitive data workflows~\citep{garza2025prvlquantifyingcapabilitiesrisks,pal2024empiricalimpactdatasanitization}. 
Traditionally, refinement pipelines have relied on heuristic rules and handcrafted scripts~\citep{lee2021surveydatacleaningmethods,li2019preprocessingmethodspipelinesdata}. Although reproducible, these pipelines become brittle as corpora, data policies, and downstream requirements evolve. The rise of LLMs enables a more flexible interface for data refinement, allowing users to specify goals in natural language across tasks such as text cleaning, quality filtering, information extraction, personally identifiable information (PII) redaction, and hallucination handling.

\input{tables/comparison}

Despite this flexibility, these operations are rarely performed in isolation. Real-world data refinement requires executing a \emph{compositional} and \emph{order-sensitive} pipeline 
% \qianli{pipeline maybe, at least "data recipe"} 
of interdependent operations over an evolving text state, where instructions dictate not only \emph{what} to apply, but also \emph{when}. Even if an LLM serves as a capable proxy for single, atomic operations, the compositional nature of refinement means that minor errors easily compound along the sequential execution, leading to eventual pipeline failure. Beyond mere composition, the order-sensitive requirement introduces an even more formidable challenge:
since each step during the pipeline execution can shift the context seen by subsequent steps, reordering operations can drastically alter intermediate states, the final edited text, and the ultimate keep/drop decision. 
% such as placing a redaction step before or after a length filter
Consequently, correctness in this setting is strictly procedural: the model must faithfully execute a latent sequence rather than merely generate a superficially plausible output.
% \qianli{you are focusing on "order-sensitive", ignoring "compositional" as a challenge. the logic can be "in multi-op case, even llm is a good proxy for single op, error can be accumulated along sequential evoking, resulting in final failure. and order-sensitive can be even more challenging"}

However, as summarized in Table~\ref{tab:benchmark_comparison}, existing benchmarks fail to evaluate this procedural faithfulness adequately. Standard instruction-driven text-editing benchmarks~\citep{editeval,fineedit} focus on isolated edits, completely overlooking compositional and order-sensitive recipes. Conversely, while recent data-centric agents tackle multi-step pipelines~\citep{datagovbench,dacomp}, their end-to-end evaluations entangle the model's intrinsic procedural understanding with sandbox execution, debugging, and code generation. To untangle these factors and isolate the underlying model's genuine data refinement capability, \emph{direct recipe execution} serves as a vital diagnostic primitive. As illustrated in Figure~\ref{fig:demo}, it directly evaluates whether an LLM can independently apply the right operations in the right order to the right intermediate states without external scaffolding. 
This critical gap raises a question: can an LLM faithfully and directly execute compositional, order-sensitive data refinement recipes?

To answer this question, we introduce \textbf{CDR-Bench}, a comprehensive benchmark designed to evaluate LLMs on compositional, order-sensitive data refinement. 
% As shown in Figure~\ref{fig:bench}, CDR-Bench comprises 3,462 tasks spanning four real-world data refinement domains: Web Refinement, LaTeX Refinement, RAG Preparation, and Privacy Redaction, covering 29 diverse operators and 63 unique recipe templates.
As shown in Figure~\ref{fig:bench}, CDR-Bench comprises 3,462 tasks across four real-world data refinement domains—Web Refinement, LaTeX Refinement, RAG Preparation, and Privacy Redaction—covering 29 operators, 63 recipe templates, two atomic tracks (Atomic-M/F) and three compositional tracks (Agnostic-M, Order-M, and Order-F).
Crucially, all tasks carry deterministic reference outputs, enabling exact, objective evaluation without relying on LLM-as-a-judge paradigms. Our evaluation framework categorizes model capabilities into three distinct levels: atomic operator execution, order-agnostic recipe execution, and order-sensitive recipe execution. To rigorously assess performance across these levels, we further propose three complementary metrics: Recipe Success (RS) for exact match verification, Order-Consistent Success (OCS) to measure robustness against order permutations, and Refinement Gain (RG) for partial progress.

% \textcolor{red}{Through extensive experiments on over 10 state-of-the-art LLMs, we uncover consistent failure patterns. Models exhibit a sharp performance degradation when shifting from atomic operations to compositional settings. Furthermore, .... Ultimately, CDR-Bench provides a reproducible diagnostic tool for compositional recipe execution, serving as a foundational primitive for both efficient refinement and future agentic data pipelines.}

Through extensive experiments on over ten state-of-the-art LLMs, we find that compositional recipe execution exposes a fundamental limitation that single-operator performance does not predict. Models frequently produce outputs that look plausible on the surface while silently violating the intended execution order, and fewer than 5\% (Order-M) and 19\% (Order-F) of order-sensitive groups are solved correctly across all tested orderings despite overall text quality remaining relatively stable. Deferred filtering decisions prove especially fragile, with exact recipe success falling by over 47 percentage points (pp) simply by moving a filter from before to after a sequence of transformations, a gap that neither prompt engineering nor few-shot demonstrations fully close. These findings suggest that progress in data refinement demands a shift from surface-level text editing to genuine procedural faithfulness, and CDR-Bench provides a reproducible testbed to measure and drive that progress. 

%% file: tables/comparison.tex
\begin{table*}[t]
\centering
\scriptsize
\setlength{\tabcolsep}{4pt}
\renewcommand{\arraystretch}{0.95}
\resizebox{\textwidth}{!}{
\begin{tabular}{l c c c c c}
\toprule
\textbf{Benchmark}
&
\textbf{Input}
&
\begin{tabular}[c]{@{}l@{}}\textbf{Primary}\\\textbf{Output}\end{tabular}
&
\begin{tabular}[c]{@{}c@{}}\textbf{Compositional}\\\textbf{Recipe}\end{tabular}
&
\begin{tabular}[c]{@{}c@{}}\textbf{Order}\\\textbf{Sensitive}\end{tabular}
&
\textbf{Evaluation Method}
\\
\midrule
\multicolumn{6}{l}{\textit{Task-Specific Editing Benchmarks}} \\

TAB~\citep{pilan-etal-2022-text}
& Text
& \begin{tabular}[c]{@{}c@{}}Edited Text\end{tabular}
& \xmark
& \xmark
& Human Annotation
\\

EditEval~\citep{editeval}
& Text
& Edited Text
& \xmark
& \xmark
& Human Annotation
\\

InstrEditBench~\citep{fineedit}
& Text
& Edited Text
& \xmark
& \xmark
& Synthetic Annotation
\\

OpenPII-1M~\citep{ai4privacy_openpii_1m_2026}
& Text
& \begin{tabular}[c]{@{}c@{}}Edited Text\end{tabular}
& \xmark
& \xmark
& Synthetic Annotation
\\

\midrule
\multicolumn{6}{l}{\textit{Data Agent Benchmarks}} \\
AutoDCWorkflow~\citep{autodcworkflow}
& Tables
& Tables, Scripts
& \xmark
& \xmark
& Execution-based
\\
DataGovBench~\citep{datagovbench}
& Tables, Text
& Scripts
& \checkmark
& \xmark
& Execution-based
\\
DCA-Bench~\citep{dcabench}
& Files
& Reports
& \xmark
& \xmark
& LLM-as-a-judge
\\
KramaBench~\citep{kramabench}
& Files
& Scripts
& \checkmark
& \xmark
& LLM, Execution-based
\\
DAComp~\citep{dacomp}
& Files
& Scripts, Reports
& \checkmark
& \xmark
& LLM, Execution-based
\\
\midrule
\textbf{CDR-Bench (Ours)}
& Text
& \begin{tabular}[c]{@{}c@{}}Decision, Edited Text\end{tabular}
& \checkmark
& \checkmark
& \begin{tabular}[c]{@{}c@{}}Deterministic Recipe\\Execution\end{tabular}
\\
\bottomrule
\end{tabular}}
% \caption{
% Scope comparison with representative task-specific editing benchmarks and data-centric agent benchmarks. Existing works evaluate either
% code-driven workflows over structured data or isolated content-level curation
% tasks. CDR-Bench instead targets compositional, order-sensitive refinement
% over unstructured text with deterministic evaluation, isolating recipe
% execution from coding and tool interaction.
% }
\caption{Scope comparison with representative task-specific editing benchmarks and data-centric agent benchmarks. CDR-Bench uniquely requires faithful, order-sensitive execution of compositional refinement recipes over raw text with deterministic evaluation. }
\label{tab:benchmark_comparison}
\vspace{-1\baselineskip}
\end{table*}

%% file: sections/method.tex
\begin{figure*}[htbp]
    \centering
    \includegraphics[width=\linewidth]{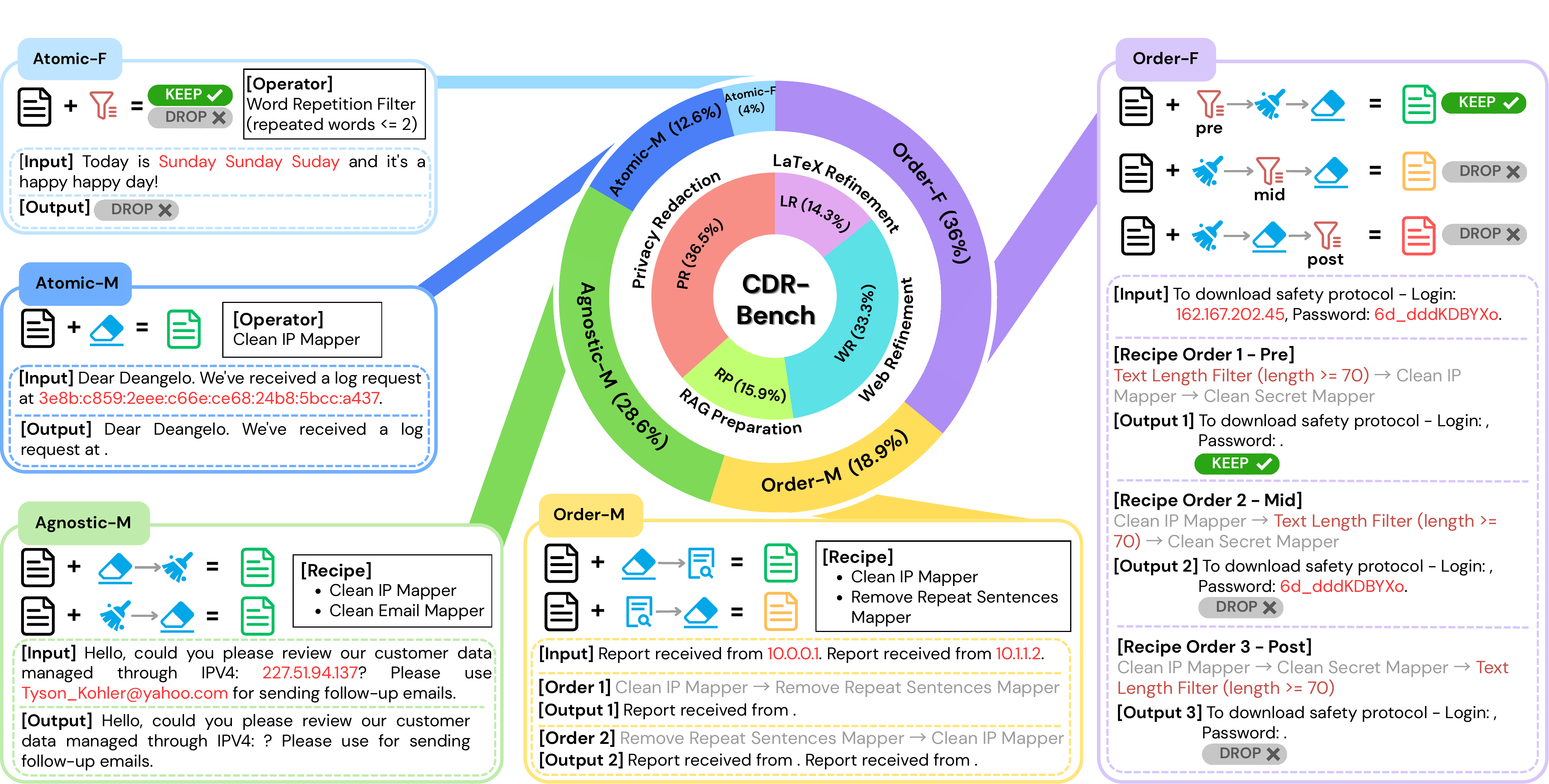}
    \caption{Overview of the CDR-Bench. The benchmark spans four data refinement domains (Privacy Redaction, Web Refinement, LaTeX Refinement, and RAG Preparation).
    The benchmark evaluates three levels of recipe execution: 
    (i) single operators (\textbf{Atomic-M/F}); 
    (ii) order-agnostic recipes (\textbf{Agnostic-M}); and 
    (iii) order-sensitive recipes: mapper permutations (\textbf{Order-M}) and filter placements across Pre/Mid/Post positions (\textbf{Order-F}).
    Here, Mappers (M) and Filters (F) denote transformation and filtering operators, respectively. Percentages are computed over unique recipe templates across domains and tracks.}
    \label{fig:bench}
    \vspace{-1\baselineskip}
\end{figure*}

\section{CDR-Bench}
In this section, we introduce the definition, dataset construction pipeline, metrics and statistics.

% CDR-Bench contains 3,145 high-quality compositional tasks and 174 auxiliary atomic tasks spanning four real-world data refinement domains. The benchmark evaluates three levels of capability: atomic operator execution, order-agnostic recipe execution, and order-sensitive recipe execution. We describe the benchmark construction pipeline below and summarize the overall benchmark composition in Table~\ref{tab:cdrbench-overview}. Detailed statistics and per-track breakdowns are provided in Appendix~\ref{appendix:cdrbench-stats}. [Yuchen TODO]

\subsection{Task Formulation}
\label{sec:task_formulation}
\paragraph{Compositional Data Refinement}
Our benchmark evaluates whether large language models can execute \emph{compositional data-refinement recipes} described in natural language. A refinement recipe $r=(o_1,o_2,\dots,o_n)$ is an ordered sequence of data-processing operators\footnote{Each $o_i$ denotes a data-processing operator. Following the taxonomy in Data-Juicer~\citep{djv1,djv2}, we consider two operator types: \emph{mappers}~(M), which rewrite text according to predefined rules, and \emph{filters}~(F), which return binary \texttt{KEEP}/\texttt{DROP} decisions.} applied to an input text $t_0$.

% \textcolor{blue}{(delete) For example, a web-document refinement recipe may sequentially remove HTML tags, filter low-quality content, and anonymize PII.
% Formally, given an input text $t_0$ and a natural-language instruction $q$ describing a recipe $r$, the model predicts an execution outcome $(\hat{s},\hat{t})$, where $\hat{s}\in\{\texttt{KEEP},\texttt{DROP}\}$ denotes the predicted execution status and $\hat{t}$ denotes the resulting refined text. The prediction is evaluated against the deterministic reference execution $E(t_0,r)=(s^\star,t^\star)$.}
Formally, the deterministic reference execution of a recipe $r$ on an input $t_0$ is denoted as $E(t_0, r) = (s^\star, t^\star)$. Here, $s^\star \in \{\texttt{KEEP}, \texttt{DROP}\}$ represents the ground-truth execution status, and $t^\star$ denotes the reference edited text. 
% Specifically, $t^\star$ is the fully refined text if $s^\star = \texttt{KEEP}$, or the intermediate text immediately prior to the rejecting filter if $s^\star = \texttt{DROP}$. 
Specifically, $t^\star$ is the final text after executing all operators if $s^\star=\texttt{KEEP}$; or the last text state immediately before the first rejecting filter is applied if $s^\star=\texttt{DROP}$.
Correspondingly, given the input text $t_0$ and a natural-language instruction $q(r)$ describing the recipe $r$, the execution by a language model $M$ is defined as $M(t_0, q(r)) = (\hat{s}, \hat{t})$. The model predicts its own execution status $\hat{s}$ and the resulting text $\hat{t}$. The core task is to evaluate whether the LLM's execution $M(t_0, q(r))$ faithfully aligns with the deterministic reference $E(t_0, r)$.

\paragraph{Order-Sensitive Recipes}

% We evaluate three settings. \textbf{Agnostic-M} evaluates recipes composed of unordered mapper combinations, testing basic sequential composition without order sensitivity. \textbf{Order-M} evaluates mapper-order sensitivity, retaining only pairs where swapping operator positions produces different outcomes. \textbf{Order-F} evaluates filter-placement sensitivity by inserting a filter at three positions (\emph{pre}, \emph{mid}, or \emph{post}) within a fixed mapper sequence, retaining only groups where at least two placements yield different outcomes.

Beyond basic composition, we further evaluate whether models correctly handle execution-order dependencies. We define a recipe group as \emph{order-sensitive} if multiple recipes share the same operator set but produce different execution outcomes under different operator ordering. Formally, for a given input text $t_0$, two recipes $r_i$ and $r_j$ formed by different permutations of the same operators are order-sensitive if their reference executions diverge, i.e., $E(t_0,r_i) \neq E(t_0,r_j)$. 
% \qianli{consider a quick example}
% \textcolor{blue}{(delete because repeated content in Track Materialization.)
% We evaluate compositional execution across three settings. \textit{Agnostic-M} tests basic sequential composition using unordered mapper combinations. \textit{Order-M} tests mapper-order sensitivity, where swapping operator positions changes the outcome. \textit{Order-F} tests filter-placement sensitivity, where inserting a filter at different \textcolor{red}{execution} positions changes the outcome. This tests whether the model correctly tracks the evolving intermediate text to make the right retention decision $s^\star$ and halt at the correct state $t^\star$.}
For instance, a \emph{text-length filter} placed before redaction mappers may yield \texttt{KEEP}, while the same filter placed after yields \texttt{DROP} as the cleaned text is shorter.

\subsection{Benchmark Construction}
\label{sub:bench_construct}

\paragraph{Overview.}
Figure~\ref{fig:bench} shows an overview of the CDR-Bench dataset.
CDR-Bench is designed to evaluate whether LLMs can faithfully execute compositional data-refinement recipes grounded in real data-processing needs. The benchmark construction pipeline proceeds in four stages: (1) collecting heterogeneous corpora and activating mapper operators on individual records; (2) mining frequent operator co-occurrence patterns to identify recipe candidates; (3) materializing tracks with deterministic references; and (4) verbalizing each recipe into diverse natural-language instructions. 
% \textcolor{blue}{(delete)As all references are produced by a deterministic operator backend, every task has a verifiable gold execution outcome $E(t_0, r) = (s^\star, t^\star)$.}

\paragraph{Data Collection and Operator Activation}
We source data from Common Crawl 2026 snapshot, arXiv preprints~\citep{tiger}, Wikipedia, GovReport~\citep{govreport}, and multiple PII datasets~\citep{docpii,ai4privacy2023pii,nemotron-pii}, organized into four functional domains: Web Refinement (WR), LaTeX Refinement (LR), RAG Preparation (RP), and Privacy Redaction (PR). For each domain, we define a set of candidate mapper and filter operators targeting common refinement requirements (detailed in Appendix~\ref{app:operator_inventory}). We then perform an operator-level activation analysis on raw records to identify which mappers produce non-trivial changes on each sample, yielding a domain-specific pool of records annotated with their activated mapper operators. This annotated pool serves as the empirical foundation for the subsequent recipe mining step, ensuring that recipe construction is grounded in naturally occurring operator patterns rather than arbitrary author-defined combinations.

% \subsection{Benchmark Construction}
% \label{sub:bench_construct}
% \paragraph{Overview.}
% The construction of CDR-Bench follows a data-driven pipeline that transforms real-world cleaning patterns into rigorously validated tasks. Our core methodology involves mining meaningful cleaning recipes from raw corpora and leveraging the deterministic nature of operator execution to produce reliable ground-truth references. These recipes are then verbalized into various natural-language styles, requiring models to execute the underlying operator logic accurately regardless of the specific phrasing used in the instruction.

% \paragraph{Data Collection and Domain Assignment.}
% CDR-Bench is constructed from a suite of heterogeneous raw corpora mirroring realistic text cleaning and sanitization requirements. By aggregating representative subsets from the Common Crawl 2026 snapshot, arXiv preprints~\citep{tiger}, Wikipedia, GovReport~\citep{govreport}, and multiple PII datasets~\citep{docpii,ai4privacy2023pii,nemotron-pii}, we establish a shared data pool for evaluation. These resources are partitioned into four functional domains, namely Web Refinement (WR), LaTeX Refinement (LR), RAG Preparation (RP), and Privacy Redaction (PR), depending on which pre-defined operator categories are activated by the noise patterns within the raw text. A comprehensive inventory of the operators defining each domain is provided in Appendix~\ref{app:operator_inventory}.

\paragraph{Recipe Mining} 
Within each annotated pool, we identify representative mapper co-occurrence patterns through a greedy coverage procedure. Candidate operator combinations are ranked by the total number of records they cover, with ties broken in favor of longer combinations. This yields a compact set of recipe family anchors that capture diverse, real-world operator compositions. We then select the most frequent exact combinations within each retained family to serve as our final recipes. Pseudocode for the mining algorithm is provided in Appendix~\ref{app:recipe_mining}.

\paragraph{Track Materialization} 
Starting from the mined mapper recipes, we materialize three recipe-level evaluation tracks. For each recipe, we pair it with raw input samples that activate all of its constituent operators, executing the pipeline to obtain the deterministic reference output $t^\star$. \textit{Agnostic-M} directly instantiates recipes from fixed mapper combinations to evaluate baseline sequential execution. \textit{Order-M} constructs order-sensitive pairs by swapping operator positions, retaining only instances where the two orderings yield divergent outcomes on the same initial text $t_0$. \textit{Order-F} dynamically inspects intermediate text states to instantiate valid filters with calibrated thresholds. It then inserts each filter at three distinct positions (\emph{pre}, \emph{mid}, and \emph{post}) within a fixed mapper sequence, retaining only those groups where at least two placements produce different final outcomes (see Appendix~\ref{app:order-f} for details). Additionally, \textit{Atomic-M} and \textit{Atomic-F} are included to evaluate single-operator tasks, serving as an isolated capability baseline.
% \paragraph{Recipe Construction and Task Instantiation.}
% Starting from the mined cleaning recipes and their activated samples, we extend each recipe with filter operators to form complete compositional recipes. During Data-Juicer replay, we inspect the intermediate text states of each cleaning recipe and use them to instantiate valid filters, including both their insertion positions and threshold parameters. We then bind each resulting clean-filter recipe to samples that activate the full recipe, and replay it to obtain the deterministic reference. This yields the multi-operator task instances used for compositional recipe execution. For order-sensitivity evaluation, we group recipes that contain the same operators but place the filter at different valid positions. We retain only groups where at least two variants yield different deterministic references, forming a diagnostic subset where filter placement materially affects the refinement outcome. We categorize these placements as \emph{pre}, \emph{mid}, and \emph{post}, corresponding to filters applied before, during, or after cleaning.

\paragraph{Instruction Verbalization} 
Once all tracks are materialized, we first build a prompt library by verbalizing each recipe into natural-language instructions across 11 prompt styles (Table~\ref{tab:prompt_styles}), keeping every instruction aligned with the operator sequence, filter semantics, and threshold values defined during materialization. An LLM-based judge screens candidate instructions for functional equivalence to the target recipe, correct preservation of execution order, and natural expression of numeric constraints without exposing code-level identifiers (Figure~\ref{fig:judge_prompt}). During evaluation, each instance is paired with three prompt variants sampled deterministically via a fixed random seed from distinct styles in the library, and wrapped in a fixed response-format template, separating phrasing variation from the standardized output contract.

\subsection{Evaluation Metrics}
\label{sub:evaluation_metrics}
% We evaluate the model's predicted execution $M(t_0, q(r)) = (\hat{s}, \hat{t})$ against the deterministic reference execution $E(t_0, r) = (s^\star, t^\star)$.
% Existing text editing benchmarks typically rely on exact
% match~\citep{squad}, edit distance~\citep{editeval,xatu,editdistance}, or
% SARI~\citep{sari}. However, these metrics are insufficient for evaluating multi-step recipes.
% Exact match provides a clear correctness signal but can
% penalize trivial formatting variations. Edit distance measures the absolute
% gap to the reference but fails to indicate whether a prediction moves the
% raw input in the correct semantic direction. Finally, SARI rewards n-gram-level rewriting
% patterns such as text simplification, while our benchmark specifically requires faithful, step-by-step execution of deterministic operator logic. 
% % Moreover, none of these metrics account
% % for the binary filter outcome or reflect compositional execution performance
% % across multi-step recipes. 
% To accurately capture strict execution correctness, partial procedural progress, and sensitivity to operator permutations, We therefore propose three task-specific
% metrics, namely Recipe Success (RS), Order-Consistent Success (OCS), and
% Refinement Gain (RG), as described below.

We evaluate $M(t_0, q(r)) = (\hat{s}, \hat{t})$ against the deterministic reference $E(t_0, r) = (s^\star, t^\star)$. Standard metrics such as exact match~\citep{squad}, edit distance~\citep{editeval,xatu,editdistance}, and SARI~\citep{sari} are insufficient here: exact match penalizes benign formatting variation, edit distance does not distinguish progress from regression, and SARI targets n-gram simplification rather than deterministic operator logic. We therefore propose three task-specific metrics: Recipe Success (RS),  Order-Consistent Success (OCS), and Refinement Gain (RG).

\paragraph{Recipe Success (RS)} measures exact recipe execution under a normalized text comparison. Let
\(f(\cdot)\) be a text normalization function that strips leading and trailing whitespace, removes empty lines, and
unifies newline characters. We define
\begin{equation}
\label{eq:recipe_success}
\mathrm{RS} = \mathbf{1}\bigl[\hat{s}=s^\star \wedge f(\hat{t})=f(t^\star)\bigr].
\end{equation}
% A prediction is successful only if it correctly predicts the execution status and perfectly reconstructs the normalized reference text. For atomic sanity checks, we report
% RS separately for mapper and filter operators. For recipe-level settings,
% we report RS overall or partitioned by \emph{pre}, \emph{mid}, and \emph{post}
% placement for the Order-F setting. We also
% report RS@K, where an instance is counted as solved if the model succeeds
% under at least one of \(K\) prompt styles.
A prediction is successful only if it correctly predicts the execution status and perfectly reconstructs the normalized reference text. We report RS@K throughout, where an instance is counted as solved if the model succeeds under at least one of $K$ prompt styles.

\paragraph{Order-Consistent Success (OCS)} is used to evaluate order-sensitive recipe groups. Each group \(\mathcal{V}\)
contains recipe variants that share the identical operator set but differ in execution sequence (e.g., permuted mapper orders or varying filter placements). OCS
requires the model to correctly solve every variant within the group simultaneously:
\begin{equation}
\label{eq:order_consistent_success}
\mathrm{OCS}(\mathcal{V}) = \mathbf{1}\!\left[\,\forall\, v\in\mathcal{V}: \ \mathrm{RS}_v=1\right],
\end{equation}
where $\mathrm{RS}_v$ denotes the Recipe Success for a specific variant $v$. For mapper-filter order groups, this necessitates succeeding on the \emph{pre},
\emph{mid}, and \emph{post} variants simultaneously. 
% We also report
% OCS@K, where a variant group is considered solved if every variant is successfully executed under at least one of its $K$ prompt styles.
We also report OCS@K, where a group is considered solved if every variant succeeds under at least one of its $K$ prompt styles.

% \qianli{it is weird you design RG as a smoothed version of RS, but no metric as a smoothed version of OCS. a trivial choice is averaged RG over V (and samples). ignore this comment if it is too difficult or time-consuming}

% \paragraph{Refinement Gain (RG)}
% Since RS and OCS do not capture partial editing progress, we introduce RG as a secondary measure of normalized edit-distance improvement toward the reference. 
\paragraph{Refinement Gain (RG)} captures partial editing progress not reflected in RS or OCS, measuring normalized edit-distance improvement toward the reference.
Let \(d(\cdot,\cdot)\) denote edit distance. We define
\begin{equation}
\mathrm{RG} = \max\!\left\{0,\,1 - \frac{d(\hat{t},t^\star)}{d(t_0,t^\star)+\epsilon}\right\},
\end{equation}
where $\epsilon>0$ is a small constant to prevent division by zero when the initial input already perfectly matches the reference. RG $=1$ when the prediction matches the reference, and $0$ when it fails to reduce the edit distance relative to the original input.
% where $\epsilon>0$ is a small constant to prevent division by zero when the initial input already perfectly matches the reference state. 
% RG equals \(1\) when the predicted text matches the reference, including
% cases where the input already satisfies the recipe. It equals \(0\) when
% the prediction fails to reduce the edit distance to the reference relative
% to the original input, assigning no partial credit to predictions that move away
% from the reference. We report RG for recipe-level settings where quantifying partial
% text progress is informative.

\subsection{Benchmark Statistics}
\label{sec:bench_status}
% CDR-Bench contains 3,145 high-quality compositional tasks and 174 atomic tasks spanning four data refinement domains. 
The core CDR-Bench benchmark contains 3,462 high-quality tasks spanning four data refinement domains, including 3,288 compositional tasks and 174 atomic tasks.
Detailed statistics are provided in Appendix~\ref{app:benchmark-statistics}.

%% file: sections/experiment.tex
\section{Experimental Results}

\input{tables/exp_main}

\subsection{Experiment Setup}
\paragraph{Evaluated Models}
We evaluate a range of state-of-the-art LLMs, spanning open-source models such as Qwen3.6~\citep{qwen3.6-27b,qwen3.6-35b-a3b}, DeepSeek-V4~\citep{deepseekv4}, GLM~\citep{glm5}, Gemma-4~\citep{gemma4}, Llama-4~\citep{llama4scout}, and Kimi-K2.6~\citep{kimik26}, as well as proprietary model families such as Claude~\citep{claude45opus,claude46opus,claude47opus,claude46sonnet} and GPT~\citep{gpt54, gpt55}. Since data refinement is often applied at corpus scale, where low-latency and cost-efficient inference is preferred, we disable reasoning modes whenever explicit controls are available. Additional model setting details are provided in Appendix~\ref{app:hyperparameters}.

% \paragraph{Prompting Strategies}
% We evaluate four prompting strategies. \textit{Direct Mode} provides the user requirement and raw input text, and asks the model to return the final tagged output directly. \textit{Few-Shot Mode} prepends two solved examples from the same evaluation track, excluding the current instance, before asking the model to solve the target case. \textit{Plan-First Mode} asks the model to first restate the requested refinement procedure as an ordered execution plan, including relevant rules or thresholds, and then produce the final output. \textit{State-Aware Mode} further encourages the model to identify intermediate text states and specify which operation or filter should be applied to each state, especially when order affects the result. Direct Mode is used as the default setting, while the other strategies test whether demonstrations, explicit planning, or intermediate-state reasoning can mitigate recipe-execution failures. Detailed prompt templates are provided in Appendix~\ref{app:eval_prompt}.

\paragraph{Prompting Strategies}
We evaluate four prompting strategies. \textit{Direct Mode} serves as the default, asking the model to produce the final output directly. \textit{Few-Shot Mode} prepends two solved examples from the same track. \textit{Plan-First Mode} asks the model to first restate the recipe as an ordered execution plan before producing the output. \textit{State-Aware Mode} further encourages the model to identify intermediate text states and specify which operation or filter applies to each. The latter three strategies test whether demonstrations, explicit planning, or intermediate-state reasoning can mitigate recipe-execution failures. Prompt templates are provided in Appendix~\ref{app:eval_prompt}.

\subsection{Main Results}
\paragraph{Performance across Different LLMs}
% (1) 多步compostional的bottleneck：atomic-m 到 agonistic-m以及order-m rs下降，rs显著下降 across all models（除了qwen3.6-35b-a3b有一个例外），说明模型处理复杂recipe的能力显著小于处理single task 数据处理任务
% (2) 细粒度中间文本状态unawareness bottleneck：agonistic-m 到order-m （order-m是调换了recipe中两个mapper的顺序结果发生改变的sample），模型rs显著下降，rg波动维持，ocs很低，甚至大部分开源模型是0，说明模型能做到大致方向的改写（往正确的改动结果靠近），但是没法有效分别中间不同state文本的fine-grained细微差异，导致ocs骤降
% (3) filter在recipe中出现的位置影响表现：在mapper/filter混合的recipe中，先进行filter若不成功则直接中止的任务最为简单（pre），而filter出现在中间和后面则需要模型针对改写后的文本进行统计，若模型无法有效预测出其改写后的文本状态，会导致准确率骤降
Table~\ref{tab:main_results} reports the main results on CDR-Bench. Although proprietary models achieve stronger absolute performance, both open- and closed-source LLMs exhibit similar degradation patterns. We highlight three key findings:

\textit{(1) Atomic-to-recipe degradation.}
Models become substantially less reliable when refinement operations are composed into multi-step recipes rather than executed in isolation. In order-agnostic settings, GPT-5.4 drops from 30.30\% RS@3 on Atomic-M to 22.74\% on Agnostic-M, while Gemma-4 (Gemma-4-31B-IT) shows a similar 4 pp decline. The degradation becomes even more severe in order-sensitive settings, where GPT-5.4 further falls to 13.99\% on Order-M and Gemma-4 drops from 25.55\% to 8.04\%. Similar trends hold across nearly all evaluated models, suggesting that procedural composition itself emerges as a major bottleneck for current LLMs.

% \textit{(2) Order sensitivity reveals fine-grained state confusion.} Beyond the general cost of composition, reordering transformations exposes a deeper failure mode. While refinement gain (RG) remains comparatively stable under different orderings, with Gemma-4 still achieving 59.32 on Order-M, exact recipe success collapses in the same setting, with DeepSeek-V4-Flash falling from 23.94\% to 4.90\% RS@3. This divergence suggests that models can produce broadly plausible refinements but fail to track the precise intermediate states induced by different operation orders. Group-level order consistency (OCS@3) peaks at only 4.90\% across all models, confirming that this failure is not occasional but systematic.

% \textit{(2) Order sensitivity exposes a gap between plausible refinement and faithful execution.}
% Changing the execution order of transformations causes exact recipe success to collapse, even when overall refinement quality remains relatively stable. For example, DeepSeek-V4-Flash drops from 23.94\% RS@3 on Agnostic-M to 4.90\% on Order-M, while RG changes much less across the two settings. This divergence suggests that models can often produce directionally correct refinements, yet fail to faithfully follow the intended execution order. Consistently, group-level OCS@3 remains extremely low across all models with the highest recorded value being only 4.90\%.

\textit{(2) Order sensitivity exposes a gap between plausible refinement and faithful execution.}
% Reordering transformations collapses exact recipe success even when overall refinement gain remains stable. DeepSeek-V4-Flash drops from 23.94\% RS@3 on Agnostic-M to 4.90\% on Order-M, while RG changes much less. This divergence suggests that models can often produce directionally correct refinements, yet fail to faithfully follow the intended execution order. Consistently, group-level OCS@3 remains extremely low across all models with the highest recorded value being only 4.90\%.
Reordering transformations collapses exact recipe success even when the overall refinement gain remains stable. DeepSeek-V4-Flash drops 19.0 pp from Agnostic-M to Order-M in RS@3 while RG changes much less, suggesting models can produce directionally correct refinements yet fail to follow the intended execution order. This pattern further manifests within Order-M as a consistent gap between recipe success and group-level order consistency, where GPT-5.4 reaches 13.99\% RS@3 but only 4.90\% OCS@3. Models can occasionally produce correct individual recipes yet consistently fail to maintain order across a full operator group.

\textit{(3) Deferred decisions become increasingly unreliable.}
% Order-F further shows that filtering decisions become substantially harder when they must be applied after multiple transformations. GPT-5.4 declines from 62.04\% RS@3 in pre-filter settings to 25.03\% and 14.97\% in mid- and post-filter settings, respectively. Qwen3.6-35B-A3B exhibits a similar pattern, dropping from 50.98\% to around 12\%. This suggests that downstream decisions become unreliable once they depend on transformed intermediate text states.
Filtering decisions degrade substantially when applied after preceding transformations. GPT-5.4 declines from 62.04\% RS@3 in pre-filter settings to 25.03\% and 14.97\% in mid- and post-filter settings, respectively. Qwen3.6-35B-A3B exhibits a similar pattern, 
%dropping from 50.98\% to around 12\%, 
indicating that downstream decisions become unreliable once they depend on transformed intermediate states.

\paragraph{Performance over Different Recipe Length}

\begin{figure*}[t]
    \centering
    \includegraphics[width=0.95\textwidth]{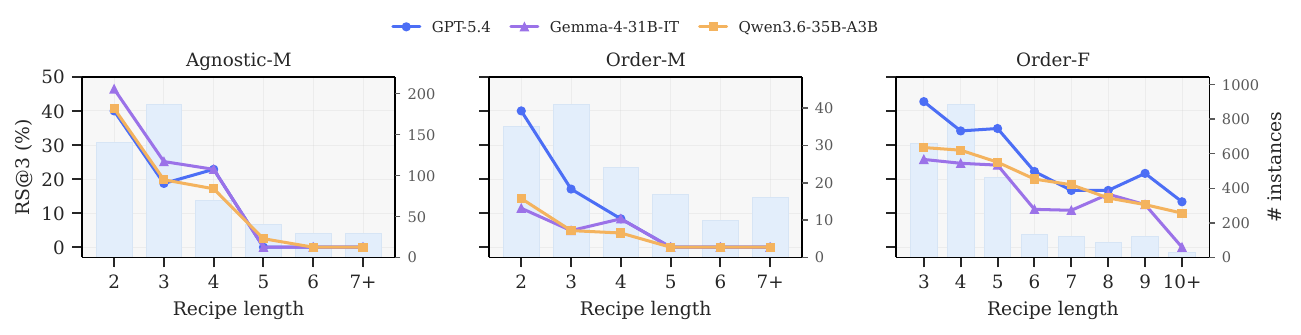}
    \caption{
    Performance over different recipe lengths on three compositional recipe tracks.
    Lines show RS@3 of representative LLMs, and light-blue bars indicate the number of instances in each recipe-length bucket.
    }
    \label{fig:recipe_length}
    \vspace{-1\baselineskip}
\end{figure*}

% Figure~\ref{fig:recipe_length} analyzes how recipe length affects compositional recipe execution. The light-blue bars indicate the number of instances in each length bucket, and the curves report RS@3 for three representative LLMs. Across all three recipe-level compositional tracks, performance consistently decreases as recipes become longer, showing that length itself is a strong impediment to faithful execution. This trend is particularly pronounced in transformation-only recipes. On both Agnostic-M and Order-M, all three models nearly collapse to zero RS@3 once the recipe contains more than five operations, suggesting that current LLMs struggle to preserve long chains of text transformations even when the individual operations are executable in isolation. The drop is sharper in the order-sensitive setting. Compared with Agnostic-M, Order-M degrades more rapidly as recipe length increases, indicating that the difficulty comes not only from longer procedural context, but also from tracking the evolving intermediate text states induced by operation order. Order-F shows a more gradual decline and remains non-zero for longer recipes. This does not necessarily mean that mixed transformation-filtering recipes are intrinsically easier. In Order-F, the transformation order is fixed and the benchmark varies where the filtering operation is applied, whereas Order-M directly perturbs the order of transformations. The latter requires models to distinguish fine-grained intermediate states, making long-horizon execution increasingly fragile as recipe length increases.

Figure~\ref{fig:recipe_length} shows that performance consistently decreases as recipes grow longer across all three compositional tracks. On Agnostic-M and Order-M, all three models nearly collapse to zero RS@3 beyond five operations. Order-M degrades more rapidly than Agnostic-M, indicating that tracking evolving intermediate states under operator reordering compounds the difficulty introduced by length alone. Order-F shows a more gradual decline, as its transformation order is fixed and only the filter placement varies, requiring less fine-grained state discrimination than Order-M.

\paragraph{Performance across Domains}

\begin{figure}[t]
    \centering
    \includegraphics[width=\columnwidth]{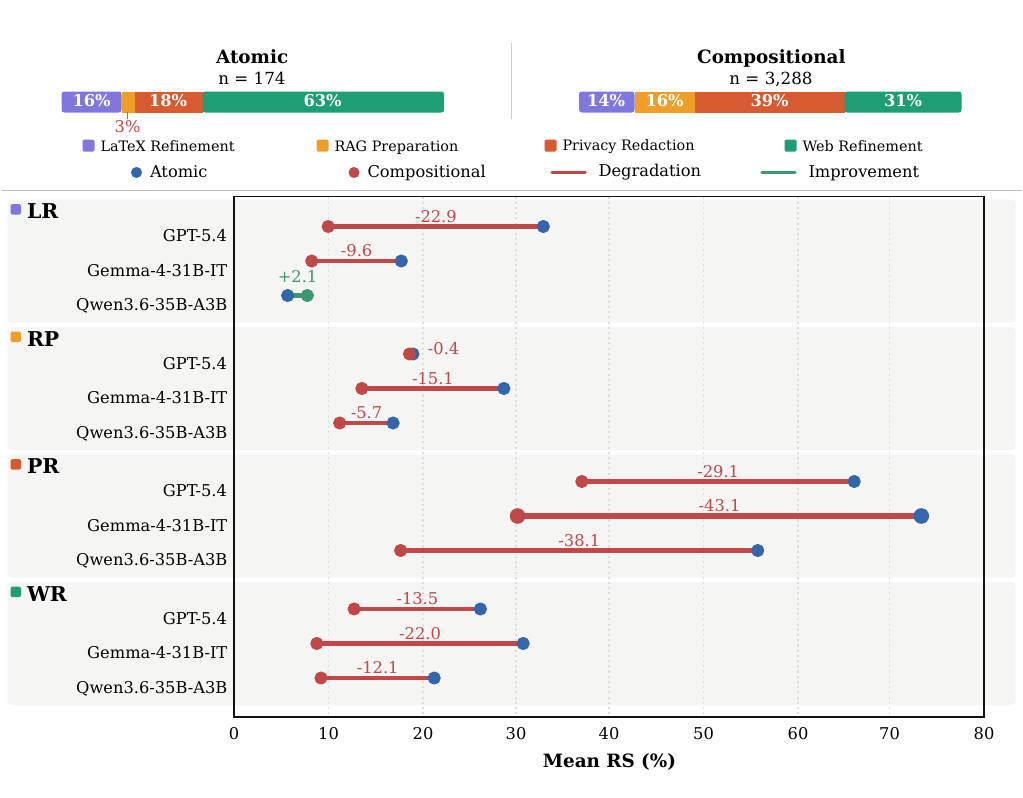}
    \caption{Atomic and compositional RS across domains, with per-model degradation gaps (pp).}
    \label{fig:domain_performance}
    \vspace{-1\baselineskip}
\end{figure}

% Figure~\ref{fig:domain_performance} shows the absolute performance and compositional degradation across domains. Privacy Redaction (PR) achieves the highest atomic RS across all models, consistent with its operators targeting localized patterns such as email addresses, IP addresses, and URLs. By contrast, Web Refinement (WR) and LaTeX Refinement (LR) score lower atomically, as their operators require reasoning over broader document context: identifying HTML structure or locating copyright boilerplate spans multiple lines and depends on document-level layout rather than isolated token patterns. Despite this, PR suffers the steepest compositional drops ($-29.1$ pp for GPT-5.4, $-38.1$ pp for Gemma-4). This inversion suggests that atomic familiarity does not confer compositional robustness, and the bottleneck shifts to multi-step state tracking regardless of operator simplicity.

Figure~\ref{fig:domain_performance} shows performance and compositional degradation across domains. PR achieves the highest atomic RS@3, as its operators target localized patterns such as emails and URLs. WR and LR score lower atomically, since their operators require reasoning over broader document structure. Despite this, PR suffers the steepest compositional drops ($-43.1$ pp for Gemma-4, $-38.1$ pp for Qwen3.6-35B-A3B), suggesting that atomic familiarity does not confer compositional robustness and the bottleneck shifts to multi-step state tracking regardless of operator simplicity.

\paragraph{Effect of Different Prompting Strategies}
Table~\ref{tab:main_results} compares prompt-level strategies against Direct Mode. \textit{Few-Shot} improves atomic execution by 5--6 pp for Qwen3.6-35B-A3B but offers little benefit at the recipe level. \textit{Plan-First} brings moderate gains on order-sensitive settings for GPT-5.4 but remains inconsistent across models. \textit{State-Aware} is the most effective strategy: by explicitly identifying intermediate text states and specifying which operation or filter applies to each, it consistently improves order-sensitive performance, increasing Order-F OCS@3 by 4 pp for GPT-5.4 and 9 pp for Qwen3.6-35B-A3B, while also benefiting Agnostic-M, suggesting that explicit state tracking helps beyond strictly order-sensitive settings. Nevertheless, absolute OCS@3 scores remain low across all strategies, indicating that prompt engineering alone is insufficient to fully resolve recipe execution failures.

\paragraph{Effect of Instruction Styles}
Figure~\ref{fig:rs_style_analysis} (left) shows that RS@K increases steadily with the number of styles attempted, as different phrasings provide complementary benefits, though the Atomic--Compositional gap persists across all $K$. The right panel shows that operation-enumerating styles such as \textit{Step-by-Step} and \textit{Brief} outperform outcome-oriented styles such as \textit{Goal-First} and \textit{Scenario Story} on transformation-heavy tracks, as the latter require models to infer intermediate procedures from the desired end state. On Order-M, all styles collapse into a narrow performance range, confirming that instruction phrasing alone cannot compensate for the fine-grained state tracking required for order-sensitive execution.

% \begin{figure*}[htbp]
%     \centering
%     \includegraphics[width=0.9\linewidth]{figs/styles_rs.pdf}
%     \caption{RS@K curves by task family (left) and mean RS across tracks and prompt styles (right).}
%     \label{fig:rs_style_analysis}
%     \vspace{-0.5\baselineskip}
% \end{figure*}

\begin{figure}[t]
    \centering
    \includegraphics[width=0.8\columnwidth]{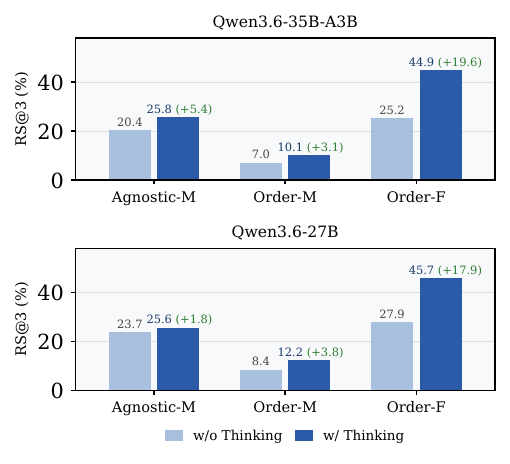}
    \vspace{-0.6\baselineskip}
    \caption{Effect of thinking mode across tracks.}
    \label{fig:thinking_ablation}
    \vspace{-0.5\baselineskip}
\end{figure}

\paragraph{Effect of Thinking Mode}
Figure~\ref{fig:thinking_ablation} shows that thinking mode substantially improves RS@3 across all compositional tracks, with the largest gains on Order-F ($+19.6$ pp for Qwen3.6-35B-A3B and $+17.9$ pp for Qwen3.6-27B), where explicit multi-step reasoning most directly benefits filter evaluation over transformed text states. Gains on Order-M are comparatively modest, as mapper ordering errors are more implicit and harder to surface even with extended reasoning.

% \paragraph{Effect of Inference-Time Strategies}
% Among prompting strategies, \textit{Few-Shot} improves atomic but not 
% compositional execution, suggesting that operator-level semantic knowledge 
% does not transfer to multi-step coordination; \textit{Plan-First} yields 
% moderate gains, indicating that explicit procedural restatement provides 
% limited but non-trivial benefit; and \textit{State-Aware Mode} is the most effective, improving Order-F OCS@3 by 4 pp for GPT-5.4 and 9 pp for Qwen3.6-35B-A3B by explicitly tracking intermediate text states, yet absolute scores remain low, indicating that prompt engineering alone cannot fully resolve recipe execution failures. Regarding instruction styles, operation-enumerating styles such as \textit{Step-by-Step} outperform outcome-oriented styles on transformation-heavy tracks, though all styles collapse into a narrow range on Order-M, confirming that phrasing cannot compensate for fine-grained state tracking either. Finally, enabling thinking mode yields the largest gains on Order-F ($+19.6$ pp for Qwen3.6-35B-A3B), where multi-step reasoning most directly benefits filter evaluation over transformed states, while gains on Order-M remain modest as mapper ordering errors are harder to surface even with extended reasoning. Full ablation results on inference-time strategies are provided in Appendix~\ref{app:ablation}.

\subsection{Error Analysis}

\begin{figure}[t]
  \centering
  \includegraphics[width=0.9\columnwidth]{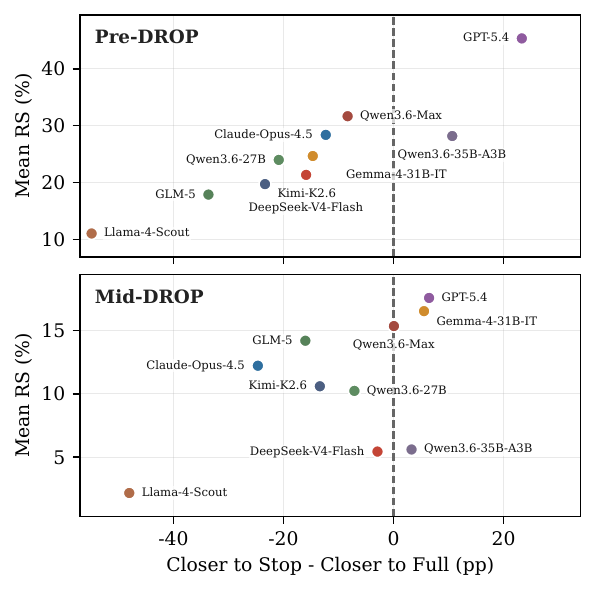}
  \caption{
  Failure to stop after a rejecting filter in Order-F DROP cases.
  Predictions are compared with the correct stopping state \(t_{\mathrm{stop}}\) and a continued-execution state \(t_{\mathrm{full}}\).
  The x-axis shows the closer-rate difference between \(t_{\mathrm{stop}}\) and \(t_{\mathrm{full}}\), and the y-axis reports mean RS.
  }
  \label{fig:order-f-drop-stop-full}
  \vspace{-0.5\baselineskip}
\end{figure}

\begin{table}[t]
\centering
\resizebox{\columnwidth}{!}{%
\begin{tabular}{lcccc}
\toprule
\textbf{Model} & \textbf{$n$} & \textbf{Closer to canonical} & \textbf{Closer to swapped} & \textbf{Tie} \\
\midrule
Gemma-4-31B-IT            & 409  & 68.0\% & 29.8\% & 2.2\% \\
GPT-5.4            & 383  & 55.4\% & 43.1\% & 1.6\% \\
Qwen3.6-35B-A3B    & 416  & 54.6\% & 43.0\% & 2.4\% \\
\midrule
\textbf{Overall}   & 1208 & \textbf{59.4\%} & \textbf{38.6\%} & \textbf{2.1\%} \\
\bottomrule
\end{tabular}}
\caption{Analysis of \textsc{Order-M} \texttt{swapped}-condition errors. Most failures are closer to the canonical ordering than to the perturbed gold, suggesting systematic resistance to instruction rather than random output.}
\label{tab:order_m_proximity}
\vspace{-1\baselineskip}
\end{table}

Figure~\ref{fig:failure-mode-analysis} summarizes the failure distribution and full definitions and breakdowns are in Appendix~\ref{app:failure-mode-analysis}. Three patterns stand out. \textit{(1) Filter-threshold errors} account for 39.3\% of failures, where models produce plausible cleaned text but assign the wrong \texttt{KEEP}/\texttt{DROP} status. \textit{(2) Deterministic execution remains brittle}: mapper tracks are dominated by formatting drift (28.4\%), missed operators (11.7\%), and under-application (9.5\%), indicating that models approximate rather than faithfully reproduce operator behavior. \textit{(3) Rejection does not reliably halt execution}: in Order-F DROP cases, models sometimes continue past a rejecting filter. As shown in Fig.~\ref{fig:order-f-drop-stop-full}, RS in Pre-DROP cases closely tracks preference for $t_{\mathrm{stop}}$ over $t_{\mathrm{full}}$, while Mid-DROP cases further require correct intermediate-state rewriting before termination.

\subsection{Real-World Scenario Evaluation}

\label{sec:real_scenario}

As a lightweight extension, we verify that the compositional difficulty identified above is not confined to rule-based operators. We evaluate four semantic tracks with human-annotated references, covering PII redaction~\citep{ai4privacy2023pii400k}, hallucination processing~\citep{fava}, safety tagging~\citep{aegis2}, and rubric scoring~\citep{helpsteer2}, each with an Atomic track over individual semantic operations and a Compositional track that requires producing all of them jointly. Results are summarized in Table~\ref{tab:semantic_pii_full}--\ref{tab:semantic_rubric_full}, with full details in Appendix~\ref{app:real_world_scenario}.

As shown in Figure~\ref{fig:semantic_atomic_comp_drop_by_domain}, the same atomic-to-compositional pattern holds across all four domains. Averaged over models and domains, atomic RS@3 is 58.8\% while compositional RS@3 is 30.8\%, an overall gap of 28.0 pp. This confirms that the composition bottleneck generalizes from deterministic operators to higher-level semantic reasoning evaluated against human ground truth, rather than being an artifact of the rule-based recipe design. The severity of the gap varies with how the subtasks are structured. It is largest when several parallel, independently meaningful decisions must all be correct under a fixed output schema, as in rubric scoring (51.1\,pp), PII redaction (29.1\,pp), and safety tagging (27.6\,pp). Hallucination processing is the exception (4.0\,pp), though not because composition is easy. Its subtasks form a coarse-to-fine progression (detection, span, type, correction) whose fine-grained stages are themselves the bottleneck, leaving little room for an additional composition penalty. Across domains, the composition bottleneck is consistent, while its severity is shaped by how the subtasks are combined.

%% file: tables/exp_main.tex
\begin{table*}[t]
\centering
\scriptsize
\setlength{\tabcolsep}{4.0pt}
\renewcommand{\arraystretch}{1.10}

\resizebox{\textwidth}{!}{%
\begin{tabular}{
@{} l
@{\hspace{6pt}} c
@{\hspace{6pt}} c
@{\hspace{7pt}} cc
@{\hspace{7pt}} ccc
@{\hspace{7pt}} ccccc
@{}}
\toprule
\textbf{Model}
&
\multicolumn{1}{c}{\textbf{Atomic-M}}
&
\multicolumn{1}{c}{\textbf{Atomic-F}}
&
\multicolumn{2}{c}{\textbf{Agnostic-M}}
&
\multicolumn{3}{c}{\textbf{Order-M}}
&
\multicolumn{5}{c}{\textbf{Order-F}}
\\
\cmidrule(lr){2-2}
\cmidrule(lr){3-3}
\cmidrule(lr){4-5}
\cmidrule(lr){6-8}
\cmidrule(lr){9-13}
&
\textbf{RS@3}$\uparrow$
&
\textbf{RS@3}$\uparrow$
&
\textbf{RS@3}$\uparrow$
&
\textbf{RG}$\uparrow$
&
\textbf{RS@3}$\uparrow$
&
\textbf{RG}$\uparrow$
&
\textbf{OCS@3}$\uparrow$
&
\textbf{RS$_{\text{pre}}$@3}$\uparrow$
&
\textbf{RS$_{\text{mid}}$@3}$\uparrow$
&
\textbf{RS$_{\text{post}}$@3}$\uparrow$
&
\textbf{RG}$\uparrow$
&
\textbf{OCS@3}$\uparrow$
\\
\midrule

\rowcolor{closedBg}
\multicolumn{13}{@{}c@{}}{\textit{Closed-Source Models}} \\
\addlinespace[1pt]

Claude Opus 4.5
& 27.27
& 71.43
& 22.13 & 49.39
& 11.54 & 51.00 & \underline{2.80}
& 42.51 & 18.56 & 13.05
& 41.17 & 8.14 \\

Claude Opus 4.6
& \underline{29.55}
& 73.81
& 22.33 & 48.06
& \underline{12.24} & 50.42 & \underline{2.80}
& \underline{57.96} & \textbf{33.17} & \textbf{23.95}
& \underline{47.03} & \textbf{18.68} \\

% Gemini-3.1-Pro
% & \textbf{33.33}
% & \textbf{97.62}
% & \textbf{26.16} & \textbf{57.23}
% & xx.xx & xx.xx & xx.xx
% & xx.xx & xx.xx & xx.xx
% & xx.xx & xx.xx \\

% Gemini-3-Flash
% & 
% & 
% &  & 
% & xx.xx & xx.xx & xx.xx
% & xx.xx & xx.xx & xx.xx
% & xx.xx & xx.xx \\

Qwen3.6-Max
& 26.15
& 76.19
& 23.94 & 50.81
& 7.34 & 48.90 & 0.70
& 46.88 & 21.82 & 17.03
& 39.44 & 8.27 \\

% Qwen3.7-max
% & 24.81
% & 73.81
% & 24.55 & 52.31
% & 9.44 & 54.88 & 1.40
% & 39.21 & 20.02 & 15.35
% & 41.22 & 7.67 \\

GPT-5.4
& \textbf{30.30}
& \textbf{88.10}
& 22.74 & \underline{53.77}
& \textbf{13.99} & \underline{52.58} & \textbf{4.90}
& \textbf{62.04} & \underline{25.03} & 14.97
& \textbf{48.47} & \underline{9.46 }\\

\quad + Few-Shot
& 31.82\dpos{1.52}
& 88.10\dpos{0.00}
& 23.38\dpos{0.64} & 52.98\dneg{0.79}
& 13.99\dpos{0.00} & 52.46\dneg{0.12} & 4.90\dpos{0.00}
& 63.37\dpos{1.33} & 25.72\dpos{0.69} & 15.18\dpos{0.21}
& 48.46\dneg{0.01} & 9.99\dpos{0.53} \\

\quad + Plan-First
& --
& --
& 23.74\dpos{1.00} & 54.45\dpos{0.68}
& 18.18\dpos{4.19} & 55.37\dpos{2.79} & 7.69\dpos{2.79}
& 62.42\dpos{0.38} & 29.38\dpos{4.35} & 17.01\dpos{2.04}
& 53.43\dpos{4.96} & 13.06\dpos{3.60} \\

\quad + State-Aware
& --
& --
& 23.84\dpos{1.10} & 54.17\dpos{0.40}
& 16.43\dpos{2.44} & 53.24\dpos{0.66} & 7.69\dpos{2.79}
& 75.15\dpos{13.11} & 29.52\dpos{4.49} & 17.07\dpos{2.10} 
& 51.18\dpos{2.71} & 13.45\dpos{3.99} \\

\midrule
\addlinespace[2pt]
\rowcolor{openBg}
\multicolumn{13}{@{}c@{}}{\textit{Open-Source Models}} \\
\addlinespace[1pt]

DeepSeek-V4-Flash
& 24.62
& 76.19
& 23.94 & 44.41
& 4.90 & 41.27 & 0.00
& 39.57 & 11.75 & 13.67
& 33.43 & 2.40 \\

Kimi-K2.6
& 29.23
& 76.19
& 24.14 & 46.36
& 9.44 & 47.77 & 0.70
& 35.25 & 16.55  & 14.39
& 35.39 & 5.88 \\

Llama-4-Scout-17B
& 16.92
& 62.50
& 13.59 & 35.95
& 3.87 & 26.66 & 0.00
& 22.41 & 4.47 & 3.99
& 26.98 & 0.86 \\

GLM-5
& 26.15
& 73.81
& \underline{24.80} & 49.42
& 10.49 & 52.05 & 2.10
& 33.93 & 22.78 & \underline{17.75}
& 37.92 & 6.47 \\

% GLM-5.1
% & 23.85
% & 78.57
% & 24.90 & 48.60
% & 9.93 & 50.29 & 2.16
% & 25.00 & 18.79 & 20.12
% & 37.09 & 5.04 \\

Gemma-4-31B-IT
& \underline{29.55}
& \underline{80.95}
& \textbf{25.55} & \textbf{56.54}
& 8.04 & \textbf{59.32} & 0.70
& 32.81 & 19.16 & 14.01
& 43.19 & 6.11 \\

% \quad + Few-Shot
% & 29.55\dpos{0.00}
% & 80.95\dpos{0.00}
% & 25.75\dpos{0.20} & 56.40\dneg{0.14}
% & 7.69\dneg{0.35} & 59.48\dpos{0.16} & 0.00\dneg{0.70}
% & 32.46\dneg{0.35} & 19.16\dpos{0.00} & 14.13\dpos{0.12}
% & 43.23 & 6.11\dpos{0.00} \\

% \quad + Plan-First
% & --
% & --
% & 25.55\dpos{0.00} & 56.26\dneg{0.28}
% & 7.34\dneg{0.70} & 57.13\dneg{2.19} & 0.00\dneg{0.70}
% & 33.89\dpos{1.08} & 19.52\dpos{0.36} & 13.41\dneg{0.60}
% & 40.52 & 4.19\dneg{1.92} \\

% \quad + State-Aware
% & --
% & --
% & 26.36\dpos{0.81} & 55.97\dneg{0.57}
% & 8.04\dpos{0.00} & 56.67\dneg{2.65} & 0.70\dpos{0.00}
% & 38.32\dpos{5.51} & 19.16\dpos{0.00} & 12.81\dneg{1.20}
% & 41.68 & 4.55\dneg{1.56} \\

Qwen3.6-27B
& 24.24
& \underline{80.95}
& 23.73 & 42.39
& 8.42 & 41.88 & 2.11
& 46.95 & 19.09 & 17.51
& 33.29 & 7.09 \\

Qwen3.6-35B-A3B
& 18.18
& 78.05
& 20.40 & 42.03
& 7.02 & 41.31 & 0.00
& 50.98 & 12.17 & 12.89
& 29.95 & 3.70 \\

\quad + Few-Shot
& 24.24\dpos{6.06}
& 83.33\dpos{5.28}
& 19.19\dneg{1.21} & 39.99\dneg{2.04}
& 6.32\dneg{0.70} & 40.41\dneg{0.90} & 0.00\dpos{0.00}
& 52.20\dpos{1.22} & 12.11\dneg{0.06} & 11.42\dneg{1.47}
& 30.41\dpos{0.46} & 2.40\dneg{1.30} \\

\quad + Plan-First
& --
& --
& 20.52\dpos{0.12} & 42.25\dpos{0.22}
& 5.94\dneg{1.08} & 40.03\dneg{1.28} & 0.00\dpos{0.00}
& 41.16\dneg{9.82} & 10.84\dneg{1.33} & 12.98\dpos{0.09}
& 31.39\dpos{1.44} & 3.55\dneg{0.15} \\

\quad + State-Aware
& --
& --
& 23.94\dpos{3.54} & 47.70\dpos{5.67}
& 9.44\dpos{2.42} & 48.52\dpos{7.21} & 2.10\dpos{2.10}
& 59.98\dpos{9.00} & 25.18\dpos{13.01} & 19.35\dpos{6.46}
& 38.61\dpos{8.66} & 12.33\dpos{8.63} \\

\bottomrule
\end{tabular}
}

\caption{%
Main results on CDR-Bench under models' non-thinking mode, evaluating baseline performances and the effects of prompt-level enhancement strategies. 
% The benchmark evaluates three levels of recipe execution: 
% (i) single operators (\textbf{Atomic-M/F}); 
% (ii) order-agnostic recipes (\textbf{Agnostic-M}); and 
% (iii) order-sensitive recipes, split into mapper permutations (\textbf{Order-M}) and filter placements across Pre/Mid/Post positions (\textbf{Order-F}).
% Here, Mappers (M) and Filters (F) represent transformation and filtering operators. 
We report recipe success (RS@3) and refinement gain (RG) across tasks, with group-level order consistency (OCS@3) additionally reported for order-sensitive recipes. 
Note that Plan-First and State-Aware strategies target recipe-level execution rather than single atomic tasks (denoted by ``--''). 
Best and second-best results among the original baseline models are \textbf{bolded} and \underline{underlined}.
}
\vspace{-1\baselineskip}
\label{tab:main_results}

\end{table*}

%% file: sections/background.tex
\section{Related Work}

\paragraph{Data Curation and Data-Centric Agent Benchmarks}
Recent work has introduced benchmarks for LLM-based data curation and data-centric agents. Some evaluate workflow construction or executable data processing over structured data, including AutoDCWorkflow and DataGovBench~\citep{autodcworkflow,datagovbench}, while broader benchmarks such as DAComp and KramaBench cover data-intelligence workflows involving SQL/Python coding, multi-source integration, data cleaning, reasoning, and report generation~\citep{dacomp,kramabench}. Other benchmarks target content-level curation tasks, including PII detection and anonymization~\citep{ai4privacy_openpii_1m_2026,pilan-etal-2022-text}, hallucination detection or correction~\citep{niu-etal-2024-ragtruth,dziri-etal-2022-faithdial}, and semantic tagging of entities and relations~\citep{wang2022mavenereunifiedlargescaledataset}.

\paragraph{LLM-Driven Data Curation and Preparation}
Recent studies have explored LLMs for data curation, cleaning, and preparation. Prior work evaluates or instruction-tunes LLMs for preprocessing operators including error detection, data imputation, schema matching, and entity matching~\citep{zhang2024largelanguagemodelsdata,zhang-etal-2024-jellyfish}. Other methods address retrieval-based repair, dependency induction for tabular cleaning, and formula synthesis for imputation~\citep{naeem2024retcleanretrievalbaseddatacleaning,biester2024llmcleancontextawaretabulardata,zhang2024sketchfillsketchguidedcodegeneration}. Systems including ChatPipe, SEED, and AutoPrep study pipeline orchestration, domain-specific curation, multi-agent table preparation, and automatic cleaning workflow generation~\citep{chen2023chatpipeorchestratingdatapreparation,chen2024seeddomainspecificdatacuration,fan2025autoprepnaturallanguagequestionaware}.

%% file: sections/conclusion.tex
\section{Conclusion}

In this paper, we introduced CDR-Bench to evaluate the faithful execution of compositional, order-sensitive data refinement recipes. 
CDR-Bench reveals a consistent gap between what LLMs appear to do and what they are actually instructed to do. Across all evaluated models, composing operators into multi-step recipes exposes failure patterns that single-operator performance does not predict: models frequently produce plausible-looking outputs while silently violating execution order, and fewer than 5\% (Order-M) and 19\% (Order-F) of order-sensitive groups are solved correctly across all tested orderings. 
% Deferred filtering proves particularly fragile, and prompt-level strategies including planning and state-aware reasoning reduce but do not close this gap. 
Real-world semantic tracks further confirm that compositional difficulty is not an artifact of rule-based operator design, though its magnitude varies with task structure. Taken together, the results point to procedural faithfulness as a distinct capability dimension that is poorly captured by existing benchmarks, and one that will need to be explicitly targeted as LLMs take on greater roles in data preparation pipelines.

% In this paper, we introduced CDR-Bench, a comprehensive benchmark designed to evaluate whether Large Language Models can faithfully execute compositional, order-sensitive data refinement recipes over evolving text states. 
% In this paper, we introduced CDR-Bench to evaluate the faithful execution of compositional, order-sensitive data refinement recipes, revealing a critical gap between what LLMs appear to do and what they are actually instructed to do. Extensive experiments demonstrate that while models succeed at isolated edits, composing operators exposes severe vulnerabilities: models frequently produce plausible-looking outputs while silently violating execution order. Furthermore, deferred filtering proves particularly fragile despite advanced prompting strategies, and real-world semantic tracks confirm this compositional bottleneck is a fundamental limitation rather than an artifact of rule-based design. Ultimately, these results highlight \emph{procedural faithfulness} as a distinct capability dimension poorly captured by existing benchmarks, establishing CDR-Bench as a vital testbed for driving reliable, multi-step data preparation pipelines.

%% file: sections/appendix.tex
\section{Benchmark Construction Details}
\subsection{Domain-Specific Operator Inventory}
\label{app:operator_inventory}

Table~\ref{tab:appendix_operator_inventory} summarizes the rule-based
operator inventory used in CDR-Bench. These operators include shared
statistical filters and domain-specific mappers for Web Refinement (WR),
LaTeX Refinement (LR), RAG Preparation (RP), and Privacy Redaction (PR).
Shared filters implement cross-domain keep/drop decisions based on
deterministic heuristic signals, such as text length, line length, and
repetition ratios. Domain-specific mappers cover common refinement
operations for crawled web text, scientific LaTeX sources, retrieval
preparation, and privacy redaction.

\input{tables/operator}

\subsection{Recipe Mining}
\label{app:recipe_mining}

\begin{algorithm}[t]
\small
\caption{Recipe Mining}
\label{alg:recipe_mining}

\begin{algorithmic}[1]

\Require Records $\mathcal{D}$, threshold $\tau$, max families $B$, max recipes $M$
\Ensure Family anchors $\mathcal{F}$, recipe sources $\mathcal{R}$

\State Extract exact mapper signatures $S_d$ from each record
\State Build exact-signature counter $C_{\text{exact}}$

\For{$l \in [L_{\min}, L_{\max}]$}
    \State Enumerate mapper subsets of size $l$
    \State Update subset counter $C_{\text{sub}}$
\EndFor

\For{each subset $A \in C_{\text{sub}}$}
    \State Compute coverage
    \[
    \mathrm{cov}(A)
    =
    \sum_{S \supseteq A}
    C_{\text{exact}}(S)
    \]
\EndFor

\State Keep candidates
$\mathcal{C}=\{A:\mathrm{cov}(A)\ge\tau\}$

\State Sort $\mathcal{C}$ by coverage,
subset size, and subset frequency

\State Initialize $\mathcal{F}\gets\emptyset$

\For{each candidate $A\in\mathcal{C}$}
    \If{$A$ covers previously uncovered signatures}
        \State Add $A$ to $\mathcal{F}$
    \EndIf

    \If{$|\mathcal{F}|=B$}
        \State \textbf{break}
    \EndIf
\EndFor

\For{each exact signature $S$}
    \State Assign $S$ to its best matching family anchor
\EndFor

\For{each family $A\in\mathcal{F}$}
    \State Select top-$M$ assigned signatures
    ranked by $C_{\text{exact}}$
\EndFor

\State \Return $\mathcal{F},\mathcal{R}$

\end{algorithmic}
\end{algorithm}

We first extract the exact mapper signature from each record and count the frequency of all observed signatures. Candidate recipe families are constructed by enumerating mapper subsets within a predefined size range and computing their coverage over exact signatures. Specifically, the coverage of a subset is defined as the total frequency of signatures containing that subset. We retain only candidates whose coverage exceeds a minimum support threshold and rank them by coverage, subset size, and subset frequency. Family anchors are then selected greedily to maximize coverage diversity, where a candidate is kept only if it covers previously uncovered signatures. Finally, each exact signature is assigned to its best matching family anchor, and the most frequent assigned signatures are retained as recipe sources.

% \paragraph{Recipe and Track Statistics.}
% Table~\ref{tab:track-stats} summarizes the benchmark composition across tracks and domains. Each order family defines a recipe family. For \textit{Order-F}, each family further expands into three recipe variants corresponding to the front, middle, and end insertion settings, whereas \textit{Order-M} and \textit{Agnostic-M} each contain a single variant per family. Evaluation rows are constructed by pairing each recipe variant with a bound sample.

% \begin{table}[t]
% \centering
% \small
% \setlength{\tabcolsep}{4pt}
% \begin{tabular}{llrrrr}
% \toprule
% \textbf{Track} & \textbf{Domain} & \textbf{Fam.} & \textbf{Rec.} & \textbf{Var.} & \textbf{Inst.} \\
% \midrule
% \multirow{4}{*}{\textit{Agnostic-M}}
% & arxiv            &  5 &  5 &   5 &  49 \\
% & knowledge\_base  &  7 &  7 &   7 &  70 \\
% & pii              & 21 & 21 &  21 & 208 \\
% & web              & 17 & 17 &  17 & 170 \\
% \midrule
% \multirow{4}{*}{\textit{Order-M}}
% & arxiv            & 13 &  8 &  13 &  32 \\
% & knowledge\_base  & 11 &  7 &  11 &  35 \\
% & pii              & 12 & 11 &  12 &  45 \\
% & web              &  9 &  7 &   9 &  31 \\
% \midrule
% \multirow{4}{*}{\textit{Order-F}}
% & arxiv            & 24 &  9 &  72 & 360 \\
% & knowledge\_base  & 25 & 10 &  75 & 375 \\
% & pii              & 65 & 23 & 195 & 975 \\
% & web              & 53 & 21 & 159 & 795 \\
% \bottomrule
% \end{tabular}
% \caption{Benchmark statistics by track and domain.}
% \label{tab:track-stats}
% \end{table}

\subsection{Filter Calibration and Insertion}
\label{app:order-f}

\begin{algorithm}[t]
\small
\caption{Filter Calibration and Insertion}
\label{alg:order-f}
\begin{algorithmic}[1]
\Require Mapper recipe $\mathcal{M} = [m_1, \ldots, m_n]$, domain filters $\mathcal{F}$, support records $\mathcal{D}$, target drop rate $\tau$
\Ensure Order-F instance groups
\For{each record $x \in \mathcal{D}$}
    \State Replay $\mathcal{M}$ on $x$, recording checkpoints $S_0, S_1, \ldots, S_n$
\EndFor
\For{each filter $f \in \mathcal{F}$}
    \For{each checkpoint $S_k$}
        \State Collect filter statistic $v_k = \mathrm{stat}(f, S_k)$ across all records
        \State Calibrate initial threshold via default percentile
    \EndFor
    \State Select \emph{front} candidate at $k=0$, \emph{end} candidate at $k=n$
    \State Select \emph{mid} candidate at $k^* = \arg\max_{0 < k < n} |\Delta \bar{v}_k|$
    \If{all three positions exist}
        \State Form order family $\{r_{\text{front}},\, r_{\text{mid}},\, r_{\text{end}}\}$
    \EndIf
\EndFor
\For{each order family}
    \State Pool statistic values $V$ across all slots and records
    \If{$f$ is \textbf{min}-type}
        \State $\theta \leftarrow \mathrm{percentile}(V, \tau)$
    \Else
        \State $\theta \leftarrow \mathrm{percentile}(V, 1-\tau)$
    \EndIf
    \For{each record $x$}
        \State $O \leftarrow [\mathrm{execute}(x, r_s, \theta) \text{ for } s \in \{\text{front}, \text{mid}, \text{end}\}]$
        \If{$|\mathrm{unique}(O)| \geq 2$}
            \State Retain group
        \EndIf
    \EndFor
    \If{retained groups $< 5$}
        \State Discard family
    \EndIf
    \State Keep at most $10$ retained groups per family
\EndFor
\end{algorithmic}
\end{algorithm}

For each mapper recipe, we replay execution and record intermediate text states $S_0,\ldots,S_n$, where $S_0$ denotes the raw input and $S_k$ denotes the state after the $k$-th mapper. Each domain filter is evaluated at every checkpoint, and an initial threshold is calibrated using the $20$th percentile for min-type filters and the $80$th percentile for max-type filters. Rather than using a fixed midpoint, the \emph{mid} position is selected as the checkpoint whose filter statistic exhibits the largest deviation from the preceding checkpoint mean, i.e., $k^\star=\arg\max_{0<k<n} |\Delta\bar{v}_k|$. During benchmark instantiation, a unified threshold $\theta$ is recalibrated by pooling statistic values across all three positions and selecting the percentile corresponding to the target drop rate ($\tau=0.5$ by default). An instance group is retained only if at least two of the three recipe variants produce different execution outcomes. Families with fewer than five retained groups are discarded.

\subsection{Recipe Verbalization Styles}

\input{tables/style}
To evaluate whether models execute the underlying recipe rather than overfitting to a single instruction wording, we verbalize each recipe using multiple user-facing styles.
Table~\ref{tab:prompt_styles} defines each style and provides representative examples. The styles vary the discourse form of the request, such as concise commands, casual requests, policy-like rules, and downstream use-case framing, while preserving the same operator sequence, execution order, and filtering semantics.

% \subsection{Judge Prompt for Instruction Validation}
% \input{figs/judge_prompt}

\subsection{Benchmark Statistics}
\label{app:benchmark-statistics}

We provide detailed statistics of CDR-Bench below.
Table~\ref{tab:benchmark-overall-stats} summarizes the overall task composition, input length distribution, recipe coverage, and instruction statistics.
Table~\ref{tab:domain-recipe-coverage} breaks down task and recipe coverage by domain, using the same domain names as in the main text: Web Refinement (WR), LaTeX Refinement (LR), RAG Preparation (RP), and Privacy Redaction (PR).

% \begin{table}[htbp]
% \centering
% \resizebox{\columnwidth}{!}{%
% \begin{tabular}{lr}
% \toprule
% \textbf{Statistics} & \textbf{Value} \\
% \midrule
% \textbf{Total Tasks} & \\
% \quad Total Tasks & 3319 (100\%) \\
% \quad Atomic Mapper Tasks & 132 (4.0\%) \\
% \quad Atomic Filter Tasks & 42 (1.3\%) \\
% \quad Order-Agnostic Mapper Tasks & 497 (15.0\%) \\
% \quad Mapper-Order Sensitivity Tasks & 143 (4.3\%) \\
% \quad Filter-Order Sensitivity Tasks & 2505 (75.5\%) \\
% \midrule
% \textbf{Input Text} & \\
% \quad Avg. Input Length (chars) & 2638.9 \\
% \quad Median Input Length (chars) & 638.0 \\
% \quad Min / Max Input Length (chars) & 27 / 9997 \\
% \quad Short / Medium / Long Inputs & 2322 / 632 / 365 \\
% \midrule
% \textbf{Recipe Coverage} & \\
% \quad Active Operators Covered & 29 \\
% \quad Unique Recipe Templates Mined & 112 \\
% % \quad Unique Recipe Templates Evaluated & 63 \\
% \quad Unique Executable Recipe Variants & 596 \\
% \quad Avg. Recipe Length Per Instance & 4.25 \\
% \midrule
% \textbf{Task Instruction} & \\
% \quad Avg. Prompt Variants Per Sample & 10.4 \\
% \quad Median Prompt Variants Per Sample & 11.0 \\
% \quad Avg. Instruction Words & 45.5 \\
% \quad Median Instruction Words & 44.0 \\
% \bottomrule
% \end{tabular}%
% }
% \caption{Overall statistics of CDR-Bench.}
% \label{tab:benchmark-overall-stats}
% \end{table}

\begin{table}[htbp]
\centering
\resizebox{\columnwidth}{!}{%
\begin{tabular}{lr}
\toprule
\textbf{Statistics} & \textbf{Value} \\
\midrule
\textbf{Total Tasks} & \\
\quad Total Tasks & 3462 (100\%) \\
\quad Atomic Mapper Tasks & 132 (3.8\%) \\
\quad Atomic Filter Tasks & 42 (1.2\%) \\
\quad Order-Agnostic Mapper Tasks & 497 (14.4\%) \\
\quad Mapper-Order Sensitivity Tasks & 286 (8.3\%) \\
\quad Filter-Order Sensitivity Tasks & 2505 (72.4\%) \\
\midrule
\textbf{Input Text} & \\
\quad Avg. Input Length (chars) & 2704.0 \\
\quad Median Input Length (chars) & 650.0 \\
\quad Min / Max Input Length (chars) & 27 / 9997 \\
\quad Short / Medium / Long Inputs & 2398 / 664 / 400 \\
\midrule
% \textbf{Recipe Coverage} & \\
% \quad Active Operators Covered & 29 \\
% \quad Unique Recipe Templates Mined & 112 \\
% % \quad Unique Recipe Templates Evaluated & 63 \\
% \quad Unique Executable Recipe Variants & 616 \\
% \quad Avg. Recipe Length Per Instance & 4.23 \\
\textbf{Recipe Coverage} & \\
\quad Active Operators Covered & 29 \\
\quad Recipe Family Anchors & 16 \\
\quad Final Executable Recipe Templates & 63 \\
\quad Materialized Executable Variants & 616 \\
\quad Avg. Recipe Length Per Instance & 4.23 \\
\midrule
\textbf{Task Instruction} & \\
\quad Avg. Prompt Variants Per Sample & 10.4 \\
\quad Median Prompt Variants Per Sample & 11.0 \\
\quad Avg. Instruction Words & 46.1 \\
\quad Median Instruction Words & 44.0 \\
\bottomrule
\end{tabular}%
}
\caption{
Overall statistics of CDR-Bench.
Family anchors are mined from operator co-occurrence patterns, executable recipe templates are selected from these families, and materialized variants count distinct executable operator sequences after order and filter-position instantiation.
}
\label{tab:benchmark-overall-stats}
\end{table}

% \begin{table}[htbp]
% \centering
% \resizebox{\columnwidth}{!}{%
% \begin{tabular}{lcccc}
% \toprule
% \textbf{Domain} & \textbf{Tasks} & \textbf{Recipes} & \textbf{Order-F} & \textbf{Order-M} \\
% \midrule
% LaTeX Refinement (LR) & 468 & 9 & 24 & 13 \\
% RAG Preparation (RP) & 486 & 10 & 25 & 11 \\
% Privacy Redaction (PR) & 1259 & 23 & 65 & 12 \\
% Web Refinement (WR) & 1106 & 21 & 53 & 9 \\
% \midrule
% Total & 3319 & 63 & 167 & 45 \\
% \bottomrule
% \end{tabular}%
% }
% \caption{
% Domain-level data and recipe coverage of CDR-Bench.
% }
% \label{tab:domain-recipe-coverage}
% \end{table}

\begin{table}[htbp]
\centering
\resizebox{\columnwidth}{!}{%
\begin{tabular}{lccc}
\toprule
\textbf{Domain} & \textbf{Tasks} & \textbf{Family Anchors} & \textbf{Recipes} \\
\midrule
LaTeX Refinement (LR) & 500 & 2 & 9 \\
RAG Preparation (RP) & 521 & 3 & 10 \\
Privacy Redaction (PR) & 1304 & 5 & 23 \\
Web Refinement (WR) & 1137 & 6 & 21 \\
\midrule
Total & 3462 & 16 & 63 \\
\bottomrule
\end{tabular}%
}
\caption{
Domain-level task and recipe coverage of CDR-Bench.
Family Anchors denotes the mined recipe-family anchors obtained from operator co-occurrence patterns; Recipes denotes the final executable recipe templates selected from these families and materialized into evaluation instances.
}
\label{tab:domain-recipe-coverage}
\end{table}

\subsection{Prompt Templates}
\label{app:eval_prompt}
We present the LLM judge prompt used for instruction validation (Figure~\ref{fig:judge_prompt}) and the four evaluation prompt templates (Figures~\ref{fig:direct_prompt}-\ref{fig:state_aware_prompt}).
\input{figs/judge_prompt}
\input{figs/eval_prompt}

\subsection{Our Position}
% \paragraph{Our Position}
Existing works primarily evaluate either code-driven data workflows over structured tables or isolated content-level curation tasks. CDR-Bench instead targets direct recipe execution for compositional, order-sensitive refinement over unstructured text with deterministic references, isolating recipe execution from coding, tool-call, and environment interaction.

\section{Experiment Details}

\subsection{Hyperparameters}
\label{app:hyperparameters}

All models are evaluated with temperature set to $0$ and a maximum output length of $32{,}768$ tokens in non-thinking mode. We use a unified output schema across all models, requiring a \texttt{status} field in \{\texttt{KEEP}, \texttt{DROP}\} and a \texttt{clean\_text} field. For RS@3 and OCS@3, each recipe is evaluated under three verbalization styles, and an instance is considered solved if any of the three prompts succeeds. Closed-source models are accessed through the DashScope API, while open-source models are deployed on $8$ NVIDIA A100 GPUs.

\begin{figure*}[htbp]
    \centering
    \includegraphics[width=\linewidth]{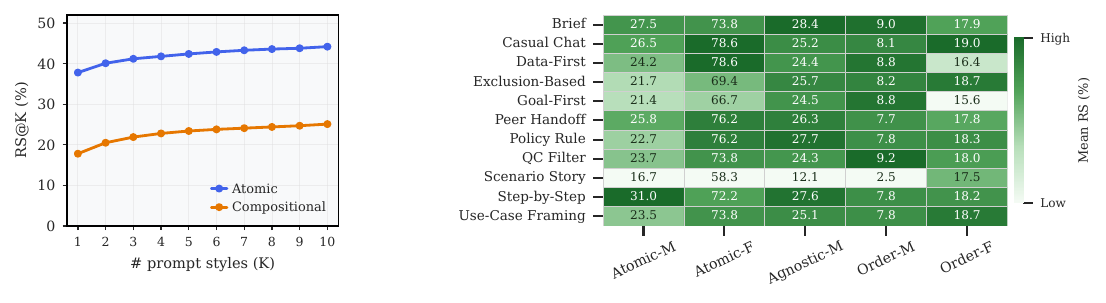}
    \caption{RS@K curves by task family (left) and mean RS across tracks and prompt styles (right).}
    \label{fig:rs_style_analysis}
    \vspace{-0.5\baselineskip}
\end{figure*}

% \begin{figure}[htbp]
%     \centering
%     \includegraphics[width=\columnwidth]{figs/thinking_ablation.pdf}
%     \vspace{-0.6\baselineskip}
%     \caption{Effect of thinking mode across tracks.}
%     \label{fig:thinking_ablation}
%     \vspace{-0.8\baselineskip}
% \end{figure}

\subsection{Failure Mode}
\label{app:failure-mode-analysis}

\begin{table}[htbp]
\centering
\resizebox{\columnwidth}{!}{%
\begin{tabular}{lp{0.62\columnwidth}}
\toprule
\textbf{Failure Mode} & \textbf{Definition} \\
\midrule
Filter Threshold Error & Incorrect \texttt{KEEP}/\texttt{DROP} decision under a filter criterion. \\
Missed Operator & One or more required operators are skipped; output remains close to input. \\
Under-application & Cleanup is partial; residual artifacts remain. \\
Over-application & Valid content is removed or modified beyond the recipe. \\
Wrong Order & Operators are applied in an order inconsistent with the recipe. \\
Formatting Drift & Output is close to reference but differs in deterministic surface formatting. \\
Semantic Rewrite & Model paraphrases or rewrites content instead of executing the recipe. \\
Instruction Misread & Model misunderstands or ignores a recipe constraint. \\
Format / Parse Error & Output cannot be parsed into the required schema. \\
\bottomrule
\end{tabular}%
}
\caption{Failure-mode taxonomy and definitions.}
\label{tab:failure-taxonomy}
\end{table}

\begin{figure*}[htbp]
    \centering
    \includegraphics[width=\textwidth]{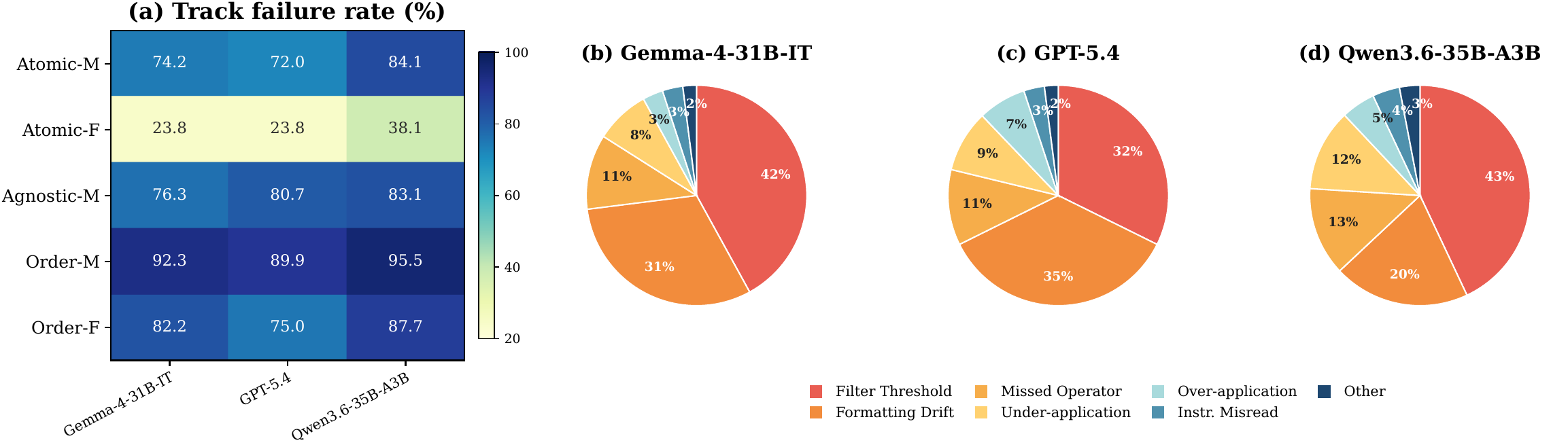}
    \caption{
    Failure mode analysis across representative models. (a) Track-level failure rates for Gemma-4-31B-IT, GPT-5.4, and Qwen3.6-35B-A3B, computed per prediction without cross-style aggregation. (b)–(d) Per-model failure mode distributions over failed predictions.
    }
    \label{fig:failure-mode-analysis}
\end{figure*}

To better understand where models fail beyond aggregate recipe-success scores, we conduct an error analysis over representative model outputs. We assign each failed prediction to one primary failure mode according to the taxonomy in Table~\ref{tab:failure-taxonomy}, separating status-level failures such as incorrect \texttt{KEEP}/\texttt{DROP} decisions from text-level execution failures such as skipped operators, incomplete cleanup, excessive deletion, order violations, formatting drift, and semantic rewriting. When multiple symptoms appear in the same prediction, we assign the most causally dominant error type, prioritizing status errors before text-level deviations. Figure~\ref{fig:failure-mode-analysis} summarizes the resulting distribution across models and tracks, with the left panel reporting track-level failure rates and the right panels showing failure mode composition per model.

\paragraph{Filter-threshold errors.} The dominant failure type (39.3\%) across all models. Models frequently produce well-formed cleaned text but misjudge the \texttt{KEEP}/\texttt{DROP} outcome, suggesting that exact numerical or statistical filtering criteria are harder to apply than the text transformations themselves.

\paragraph{Mapper execution failures.} On transformation-heavy tracks, formatting drift (28.4\%) is the leading error, followed by missed operators (11.7\%) and under-application (9.5\%). Together these indicate that models tend to approximate the requested cleanup rather than reproduce deterministic operator behavior exactly. Over-application and semantic rewrites appear less frequently but reflect a distinct failure mode where models modify content beyond what the recipe specifies. We further analyze \textsc{Order-M} errors by measuring edit distance from each incorrect model output to both the canonical (original) gold and the perturbed (swapped) gold, to determine whether failures reflect random generation or systematic order confusion.

% \begin{table}[htbp]
% \centering
% \resizebox{\columnwidth}{!}{%
% \begin{tabular}{lcccc}
% \toprule
% \textbf{Model} & \textbf{$n$} & \textbf{Closer to canonical} & \textbf{Closer to swapped} & \textbf{Tie} \\
% \midrule
% Gemma-4-31B-IT            & 409  & 68.0\% & 29.8\% & 2.2\% \\
% GPT-5.4            & 383  & 55.4\% & 43.1\% & 1.6\% \\
% Qwen3.6-35B-A3B    & 416  & 54.6\% & 43.0\% & 2.4\% \\
% \midrule
% \textbf{Overall}   & 1208 & \textbf{59.4\%} & \textbf{38.6\%} & \textbf{2.1\%} \\
% \bottomrule
% \end{tabular}}
% \caption{Analysis of \textsc{Order-M} \texttt{swapped}-condition errors. Most failures are closer to the canonical ordering than to the perturbed gold, suggesting systematic resistance to instruction rather than random output.}
% \label{tab:order_m_proximity}
% \end{table}

As shown in Table~\ref{tab:order_m_proximity}, 59.4\% of errors are closer to the canonical gold than to the required swapped ordering, suggesting that models revert to a familiar argument structure rather than following the specified permutation. The effect is strongest in Gemma-4-31B-IT (68.0\%), while GPT-5.4 and Qwen3.6-35B-A3B are more balanced, indicating partial instruction-following that nonetheless fails to align all relational slots correctly.

\paragraph{Stopping behavior in DROP cases.} Figure~\ref{fig:order-f-drop-stop-full} analyzes Order-F DROP cases, where a rejecting filter should halt execution. In Pre-DROP cases, mean RS is nearly monotonically aligned with whether predictions are closer to $t_{\mathrm{stop}}$ than $t_{\mathrm{full}}$, confirming that the primary error is continuing execution after rejection. GPT-5.4 achieves the highest RS together with the strongest preference for $t_{\mathrm{stop}}$. Smaller models such as Qwen3.6-35B-A3B obtain lower overall RS but remain relatively close to $t_{\mathrm{stop}}$, showing stronger stopping behavior than their overall execution performance would suggest. Mid-DROP cases impose an additional requirement: models must correctly rewrite the intermediate text state before terminating, making rewriting quality a further bottleneck for smaller open models. Gemma-4-31B-IT stands out among similarly sized models with relatively high RS and stronger stopping behavior, reflecting both better intermediate-state rewriting and more reliable early termination.

\section{Real-Scenario Evaluation Details}
\label{app:real_world_scenario}

\subsection{Motivation and Design}

The main evaluation tracks rely on rule-based Data-Juicer operators, which provide deterministic references but naturally invite the question of whether the observed trends extend beyond operator replay. To complement the main benchmark, we therefore evaluate four semantic-extension domains grounded in realistic data-processing scenarios. The goal is not to recreate CDR-Bench in a second form, but to test whether its central observations remain relevant in settings with human annotations, higher-level semantic judgments, and less explicitly procedural task definitions.

These tracks differ from the main evaluation in three key respects. (1) Their operations correspond to semantic tasks such as entity recognition, hallucination handling, semantic category tagging, and scoring under semantic rubrics, rather than deterministic rule execution. (2) Ground truth comes from human-annotated datasets rather than operator replay. (3) The tasks involve higher-level semantic reasoning that is only weakly exercised in rule-based tracks. We therefore treat this section as a targeted real-world scenario study that complements, rather than replaces, the main benchmark.

Note that the model pool in this section differs slightly from the main benchmark. Since this extension is designed to test whether the atomic-to-compositional trend generalizes, rather than to compare models head-to-head, we evaluate a refreshed set of more recent models. The persistence of the same gap under a newer and stronger model pool further indicates that the bottleneck is not an artifact of any particular model generation.

\subsection{Semantic Domains}

For each semantic domain, we construct an \emph{Atomic} track that evaluates individual semantic operations and a \emph{Compositional} track that requires combining the corresponding operations in a single output. Unless otherwise specified, we report the same RS@3 convention used in the main benchmark. Each instance is prompted with three fixed styles: direct, imperative checklist, and application context. A prediction is counted as successful if any of the three prompted outputs matches the reference under the task-specific scorer.

For text-producing tasks, such as PII redaction and hallucination correction, we use normalized exact match over the required output text. For structured-output tasks, such as rubric scoring and safety tagging, we require exact matching of the requested JSON fields. Atomic scores are computed separately for each semantic subtask and then macro-averaged within a domain. Compositional scores are computed on the paired full-output task. We define the composition gap uniformly as
\begin{equation}
\label{eq:semantic_gap}
\mathrm{Gap}
=
\mathrm{RS@3}_{\mathrm{atom}}
-
\mathrm{RS@3}_{\mathrm{comp}}.
\end{equation}
This mirrors the main benchmark's atomic-to-compositional comparison while keeping each semantic operation equally weighted within its domain.

\paragraph{PII Semantic Redaction}
We use the AI4Privacy PII masking dataset\footnote{We use AI4Privacy only as an evaluation benchmark for non-commercial academic research. We sample a subset from the publicly available Hugging Face release and use it exclusively for experiments in this paper. We do not redistribute the original dataset or any derivative data release.}~\citep{ai4privacy2023pii400k}. We select 500 English samples with at least two distinct PII category groups through stratified sampling. We aggregate the original PII labels into five semantic groups: \emph{person} (names, usernames), \emph{location} (cities, streets, zip codes), \emph{contact} (emails, phone numbers), \emph{identification} (ID cards, bank accounts, tax IDs), and \emph{temporal} (dates of birth and other time-related sensitive fields in the dataset grouping). The \emph{Atomic} track evaluates redaction for each group in isolation using \texttt{[LABEL\_N]} placeholders, while the \emph{Compositional} track requires redacting all present groups simultaneously. Ground truth is the programmatic redaction output derived from dataset annotations.

\paragraph{Hallucination Detection and Correction}
We use a balanced subset adapted from FAVA~\citep{fava} with 300 samples. The dataset provides LLM-generated text annotated with hallucination spans and corrections across several types, including entity errors, relation errors, subjective injections, contradictions, inventions, and unverifiable claims. The \emph{Atomic} track decomposes hallucination processing into four subtasks: (1)~\emph{Detection}, which determines whether the text contains hallucinations; (2)~\emph{Span Extraction}, which identifies hallucinated spans; (3)~\emph{Type Classification}, which classifies the hallucination types present; and (4)~\emph{Correction}, which produces corrected text with hallucinated content removed or replaced. The \emph{Compositional} track requires executing the full processing pipeline in a single pass. We evaluate structured fields with exact JSON-field matching and correction outputs with normalized text exact match.

\paragraph{Safety Tagging}
We use Aegis-AI-Content-Safety-Dataset-2.0~\citep{aegis2}, a content-safety moderation dataset with user prompts, assistant responses, binary safety labels, and violated-category annotations. We sample 300 prompt-response pairs from the test split after filtering to examples with both prompt and response labels, balancing the sample as evenly as possible across observed prompt-response label pairs. The \emph{Atomic} track decomposes safety tagging into three subtasks: (1)~\emph{Prompt Label}, which classifies the user prompt as \texttt{safe} or \texttt{unsafe}; (2)~\emph{Response Label}, which classifies the assistant response as \texttt{safe} or \texttt{unsafe}; and (3)~\emph{Violated Categories}, which identifies the relevant safety-risk categories. The \emph{Compositional} track requires producing all three fields jointly in one JSON object, testing whether models can compose related moderation decisions into a consistent structured output.

\paragraph{Rubric Scoring}
We use HelpSteer2~\citep{helpsteer2}, a response-quality dataset of prompt-response pairs annotated along five rubric dimensions: \emph{helpfulness}, \emph{correctness}, \emph{coherence}, \emph{complexity}, and \emph{verbosity}. Each dimension is labeled on a 0--4 ordinal scale. We sample 300 prompt-response pairs from the validation split with a fixed random seed. The \emph{Atomic} track evaluates one rubric dimension at a time, requiring the model to output a JSON object containing only the requested score. The \emph{Compositional} track requires predicting all five rubric scores jointly in a single JSON object.

\subsection{Results and Analysis}

\input{tables/real_scenario_appendix}

\begin{figure*}[t]
    \centering
    \includegraphics[width=.48\textwidth]{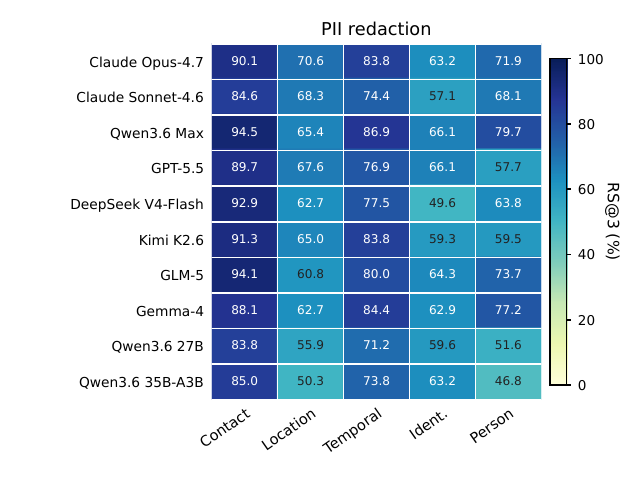}
    \includegraphics[width=.48\textwidth]{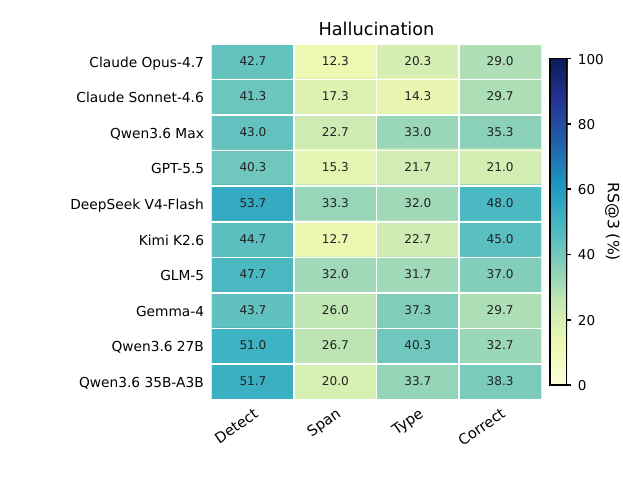}
    \includegraphics[width=.48\textwidth]{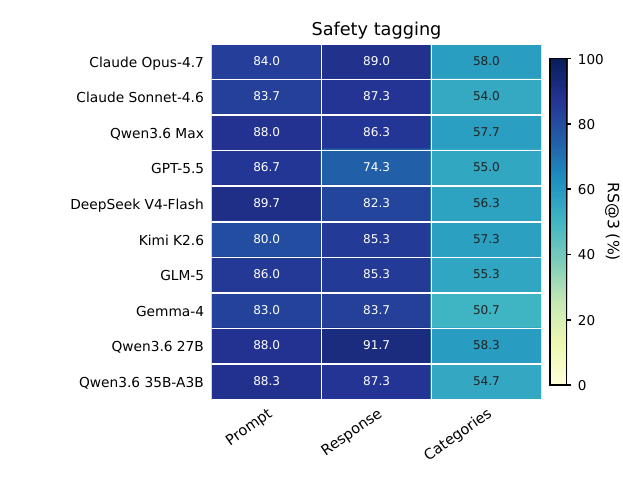}
    \includegraphics[width=.48\textwidth]{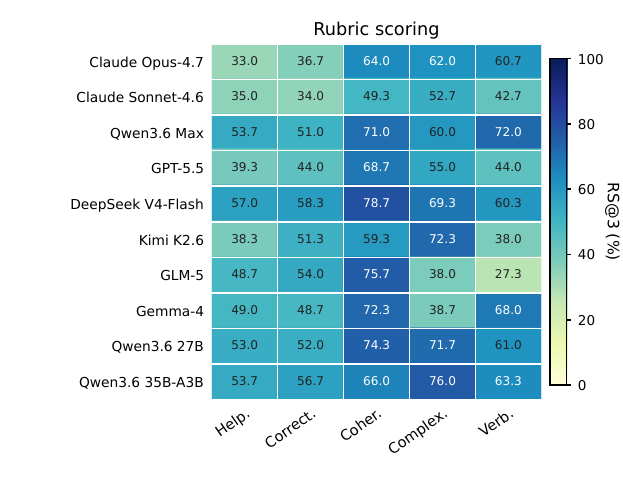}
    \caption{Atomic subtask RS@3 heatmaps for the four semantic domains. The heatmaps show that atomic difficulty is itself structured: PII varies by entity group, hallucination follows a coarse-to-fine hierarchy, rubric scoring varies by judgment dimension, and safety tagging separates easier binary labels from harder violated-category prediction.}
    \label{fig:semantic_atomic_heatmaps}
\end{figure*}

Tables~\ref{tab:semantic_pii_full}--\ref{tab:semantic_rubric_full} and Figure~\ref{fig:semantic_atomic_heatmaps} report the full results of the semantic extension. Across all four domains, models are markedly more reliable on isolated semantic operations than on the corresponding compositional task: averaged over models and domains, compositional RS@3 falls to 30.8\% against an atomic average of 58.8\%.

\begin{figure}[t]
    \centering
    \includegraphics[width=\columnwidth]{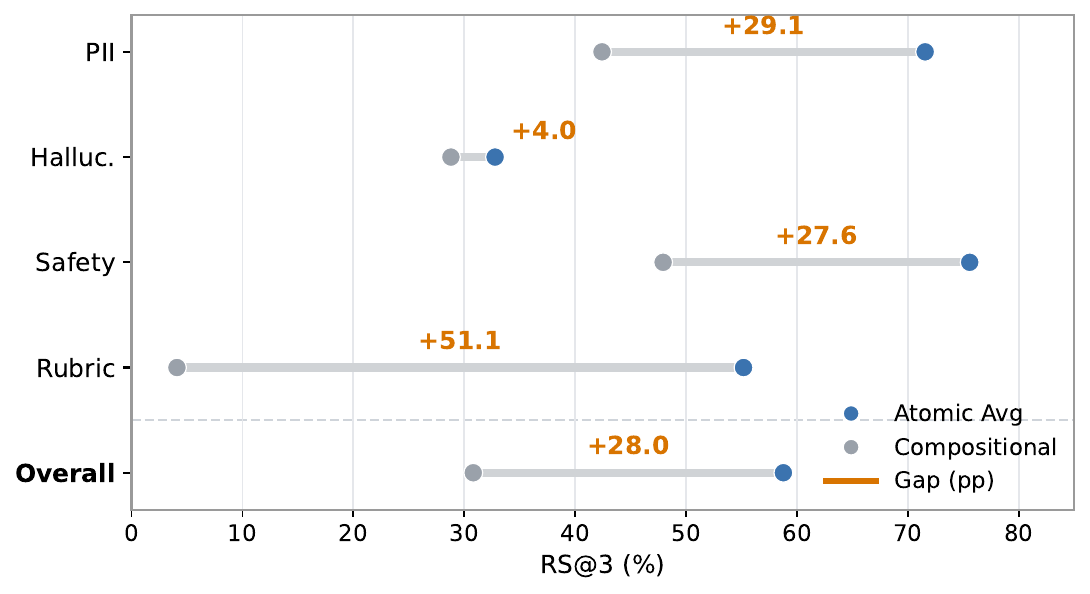}
    \caption{Atomic-to-compositional performance drop across the four semantic domains. Blue points show Atomic Avg RS@3 macro-averaged over atomic subtasks and models, gray points show Compositional RS@3, and orange labels report the gap.}
    \label{fig:semantic_atomic_comp_drop_by_domain}
\end{figure}

\paragraph{PII redaction.}
Contact information is the easiest atomic group across models, whereas identification, location, and person fields are consistently the hardest, reflecting the greater contextual reasoning they require. Composition then imposes a clear additional cost, with gaps ranging from 19.8 to 38.8 percentage points: strong per-group redaction does not guarantee that a model can satisfy all groups simultaneously in one exact output. This mirrors the transformation setting of the main benchmark, where a model must apply several localized edits while preserving all unaffected content verbatim.

\paragraph{Hallucination processing.}
Unlike the other domains, whose atomic subtasks are largely parallel, hallucination processing decomposes into a sequence of increasingly fine-grained operations: detecting whether an error is present (coarsest), localizing the offending span, classifying its type, and producing a correction (finest). The subtask scores follow this granularity ordering, with detection as the ceiling and span localization, type classification, and correction all markedly lower. The compositional score still falls below the atomic average for nearly all models, but the additional gap is comparatively small ($-1.1$ to 8.1 pp): because the fine-grained subtasks are already hard, combining them adds little on top of an already low base. The dominant difficulty in this domain is thus the fine-grained processing itself rather than the act of merging outputs.

\paragraph{Safety tagging.}
Binary safety labels are far easier than fine-grained category tagging. Prompt- and response-label classification frequently exceeds 80\%, whereas violated-category tagging is the clear bottleneck, and the compositional task requiring all three fields jointly yields gaps of 24.9 to 32.3 percentage points. Failures here are not about distinguishing safe from unsafe content, but about producing a complete, internally consistent annotation that also names the correct risk categories.

\paragraph{Rubric scoring.}
Rubric scoring shows by far the largest atomic-to-compositional degradation, with gaps of 40.1 to 61.1 percentage points. Several systems reach atomic averages above 60\% on individual HelpSteer2 dimensions, with coherence and complexity being the easiest, yet all-dimension scoring nearly collapses, with compositional scores in the low single digits. Notably, the degradation tends to be largest for the models that are strongest atomically, indicating a structural rather than knowledge-based bottleneck: a model may assign any single ordinal score correctly, but exact success requires all five judgments to be simultaneously correct within one JSON object.

\paragraph{Domain-dependent composition mechanisms.}
Taken together, these results show that semantic composition is not a single, uniform phenomenon, but reflects how the underlying subtasks are structured. PII redaction and safety tagging degrade because several parallel local constraints must be satisfied at once. Rubric scoring exhibits a stronger structured-output effect, in which exact agreement on all fields is far harder than scoring one dimension in isolation, so much so that higher atomic competence is associated with a larger compositional collapse. Hallucination processing instead spans a coarse-to-fine range of subtasks (detection $\rightarrow$ span $\rightarrow$ type $\rightarrow$ correction), where the difficulty is concentrated in the fine-grained operations and composition adds only a small further penalty. Compositional failure is thus driven not merely by the number of operations, but by whether the subtasks are parallel or vary in granularity, by the output schema, and by the granularity of the semantic judgments being combined.

% \subsection{Key Findings}

% \begin{itemize}[leftmargin=*]
%     \item \textbf{Some compositional difficulties extend to semantic settings, but not uniformly.} PII semantic redaction shows a large atomic-to-compositional drop (21--34pp), indicating that the composition bottleneck observed in the main benchmark is also relevant in semantic redaction tasks. Hallucination tasks, however, display a much smaller difference, showing that semantic composition behavior depends strongly on task structure.

%     \item \textbf{Different composition mechanisms create different challenges.} PII compositional difficulty arises from compounded failure across multiple required redaction groups, while hallucination difficulty is dominated by the correction operation itself.

%     \item \textbf{Difficulty reflects contextual reasoning demands.} PII group difficulty follows a stable gradient from format-driven categories (contact) to more context-dependent ones (person and identification-related fields), consistent across models.

%     \item \textbf{Real-world semantic tasks show heterogeneous failure patterns.} The contrast between PII and hallucination suggests that the main benchmark is most directly predictive for scenarios that require multiple precise local edits to be coordinated within one final output.
% \end{itemize}

\subsection{Key Findings}

\begin{itemize}[leftmargin=*]
    \item \textbf{Semantic composition exposes a broad reliability gap beyond the main synthetic tracks.}
    Across the four semantic domains, models are consistently better at isolated
    semantic operations than at the corresponding composed task. Averaged over
    models and domains, Atomic Avg RS@3 is 58.8\%, while Compositional RS@3 is
    30.8\%, giving an overall semantic composition gap of 28.0 percentage points.
    This indicates that the core CDR-Bench observation extends to adapted
    real-world semantic tasks.

    \item \textbf{The gap is governed by task structure rather than domain label alone.}
    Large gaps appear when success requires several independently meaningful
    decisions to be correct in one final output, as in PII redaction, safety
    tagging, and especially rubric scoring. In contrast, hallucination processing
    has a much smaller additional gap because the hardest atomic operations are
    already fine-grained diagnosis and correction.

    \item \textbf{Exact structured composition is especially brittle.}
    Rubric scoring and safety tagging show that models can often make individual
    judgments, but fail when several fields must be jointly correct under a fixed
    output schema. This suggests that semantic competence on single labels or
    dimensions does not directly translate into reliable multi-field annotation.

    \item \textbf{Atomic performance alone can overstate deployable capability.}
    Several domains contain atomic subtasks with high RS@3, yet their
    compositional scores remain much lower. Reporting only isolated operations
    would therefore obscure the practical failure mode that appears when a model
    must satisfy all requirements simultaneously.
\end{itemize}
Overall, the semantic extension supports the same central message as the main
benchmark: reliable data refinement requires not only recognizing individual
operations, but executing all required operations jointly under a strict output
contract.

\subsection{Prompt Templates}
\label{app:semantic_prompts}

We use three prompt styles per semantic-extension task (\emph{direct}, \emph{imperative checklist}, and \emph{application context}), with the direct style used as the default example in this appendix. The styles differ only in presentation: for a fixed instance, they preserve the same requested operation, input fields, and output schema. All prompts share a common system instruction that asks the model to follow the requested refinement or labeling operation exactly and to return only the specified output format.

For tagged-text tasks, including PII redaction and hallucination correction, the model returns \texttt{<status>} and \texttt{<clean\_text>} fields. For structured tasks, including hallucination detection, hallucination span extraction, hallucination type classification, rubric scoring, and safety tagging, the model returns a JSON object with the requested fields. Rubric-scoring prompts define the HelpSteer2 dimensions in the prompt and require integer scores on the 0--4 scale. Safety-tagging prompts list the Aegis safety categories and require \texttt{prompt\_label}, \texttt{response\_label}, and/or \texttt{violated\_categories} depending on whether the task is atomic or compositional. Representative prompt wrappers and domain examples are shown in Figure~\ref{fig:semantic_prompt_wrappers}--\ref{fig:semantic_rubric_prompts}.

\input{figs/semantic_prompts}

%% file: tables/operator.tex
\begin{table*}[htbp]
\centering
\scriptsize
\setlength{\tabcolsep}{4pt}
\renewcommand{\arraystretch}{1.12}

\begin{tabularx}{\textwidth}{
@{}
>{\raggedright\arraybackslash}p{0.40\textwidth}
>{\raggedright\arraybackslash}X
@{}
}
\toprule
\textbf{Operator} & \textbf{Behavior} \\
\midrule

\rowcolor{sharedBg}
\multicolumn{2}{@{}c@{}}{\textit{Shared Operators}} \\
\addlinespace[1pt]

\texttt{alphanumeric\_filter}
& Keeps text only when its alphanumeric-content ratio falls within a configured range. \\

\texttt{average\_line\_length\_filter}
& Keeps text only when the average line length falls within a configured range. \\

\texttt{character\_repetition\_filter}
& Keeps text only when character-level repetition remains within a configured range. \\

\texttt{maximum\_line\_length\_filter}
& Keeps text only when the longest line falls within a configured range. \\

\texttt{text\_length\_filter}
& Keeps text only when total text length falls within a configured range. \\

\texttt{word\_repetition\_filter}
& Keeps text only when word-level repetition remains within a configured range. \\

\texttt{words\_num\_filter}
& Keeps text only when the total word count falls within a configured range. \\

\midrule
\rowcolor{wrBg}
\multicolumn{2}{@{}c@{}}{\textit{Web Refinement (WR)}} \\
\addlinespace[1pt]

\texttt{clean\_copyright\_mapper}
& Removes copyright notices or related boilerplate from the beginning of a web text sample. \\

\texttt{clean\_html\_mapper}
& Cleans HTML markup and converts HTML content into readable plain text. \\

\texttt{clean\_links\_mapper}
& Removes HTTP, HTTPS, or FTP links from a crawled web page. \\

\texttt{extract\_tables\_from\_html\_mapper}
& Extracts table content from HTML into structured text. \\

\texttt{fix\_unicode\_mapper}
& Repairs malformed Unicode text, e.g., fixes mojibake or broken encoding. \\

\texttt{punctuation\_normalization\_mapper}
& Normalizes non-standard Unicode punctuation into standard English forms. \\

\texttt{remove\_specific\_chars\_mapper}
& Removes configured noisy symbols or visually irrelevant special characters. \\

\texttt{remove\_words\_with\_incorrect\_substrings\_mapper}
& Drops tokens containing URL-like substrings such as \texttt{http}, \texttt{www}, or \texttt{.com}. \\

\texttt{whitespace\_normalization\_mapper}
& Normalizes irregular tabs, repeated spaces, or broken spacing patterns. \\

\midrule
\rowcolor{lrBg}
\multicolumn{2}{@{}c@{}}{\textit{LaTeX Refinement (LR)}} \\
\addlinespace[1pt]

\texttt{clean\_copyright\_mapper}
& Removes copyright-related boilerplate from LaTeX source text. \\

\texttt{expand\_macro\_mapper}
& Expands macro definitions used in LaTeX source into their literal expansions. \\

\texttt{fix\_unicode\_mapper}
& Repairs malformed Unicode text in extracted LaTeX content. \\

\begin{tabular}[t]{@{}l@{}}
\texttt{latex\_figure\_context\_extractor\_mapper}
\end{tabular}
& Extracts figure-related text, captions, and nearby citing context from LaTeX source. \\

\texttt{punctuation\_normalization\_mapper}
& Normalizes irregular Unicode punctuation into common English punctuation. \\

\texttt{remove\_bibliography\_mapper}
& Removes bibliography sections at the end of LaTeX documents. \\

\texttt{remove\_comments\_mapper}
& Removes comment lines or inline comments beginning with \texttt{\%}. \\

\texttt{remove\_header\_mapper}
& Removes LaTeX preamble or header material before the main document body. \\

\midrule
\rowcolor{rpBg}
\multicolumn{2}{@{}c@{}}{\textit{RAG Preparation (RP)}} \\
\addlinespace[1pt]

\texttt{fix\_unicode\_mapper}
& Repairs corrupted characters that reduce document readability and retrievability. \\

\texttt{remove\_long\_words\_mapper}
& Removes abnormally long tokens that are unlikely to be useful for indexing. \\

\texttt{remove\_repeat\_sentences\_mapper}
& Removes repeated sentences while preserving the remaining text order. \\

\midrule
\rowcolor{prBg}
\multicolumn{2}{@{}c@{}}{\textit{Privacy Redaction (PR)}} \\
\addlinespace[1pt]

\texttt{clean\_email\_mapper}
& Removes email addresses from the text. \\

\texttt{clean\_ip\_mapper}
& Removes IPv4 and IPv6 addresses from logs. \\

\texttt{clean\_links\_mapper}
& Removes web or file links that may expose sensitive destinations. \\

\texttt{clean\_mac\_mapper}
& Removes MAC addresses from device records. \\

\texttt{clean\_path\_mapper}
& Removes file-system paths, including Unix, Windows, and UNC paths. \\

\texttt{clean\_phone\_mapper}
& Removes phone numbers from the content. \\

\texttt{clean\_secret\_mapper}
& Removes API keys, tokens, passwords, and authorization credentials. \\

% \begin{tabular}[t]{@{}l@{}}
% \texttt{remove\_words\_with\_incorrect} \\
% \texttt{\_substrings\_mapper}
% \end{tabular}
\texttt{remove\_words\_with\_incorrect\_substrings\_mapper}
& Drops words containing configured substrings like \texttt{href} or \texttt{.com}. \\

\bottomrule
\end{tabularx}
\caption{%
Operator inventory used in CDR-Bench. Operators are grouped by
domain and functional role. Shared operators are deterministic statistical
filters used across domains, while the remaining operators are
domain-specific mappers for Web Refinement (WR), LaTeX Refinement (LR), RAG
Preparation (RP), and Privacy Redaction (PR).
}
\label{tab:appendix_operator_inventory}
\end{table*}

% \begin{table}[t]
% \centering
% \small
% \setlength{\tabcolsep}{4pt}
% \renewcommand{\arraystretch}{1.10}
% \begin{tabular}{p{0.28\linewidth} p{0.62\linewidth}}
% \toprule
% \textbf{Operator Type} & \textbf{Behavior} \\
% \midrule
% Semantic PII Redaction
% & Identifies and removes sensitive information that requires contextual
% understanding rather than surface-pattern matching alone. \\

% Semantic Hallucination Detection
% & Filters text containing unsupported, inconsistent, or unfaithful content
% based on semantic consistency checks. \\
% \bottomrule
% \end{tabular}
% \caption{Semantic operator inventory used in CDR-Bench. These operators
% target refinement decisions that require contextual understanding beyond
% rule-based cleanup.}
% \label{tab:appendix_semantic_operators}
% \end{table}

%% file: tables/style.tex
\begin{table*}[htbp]
\centering
\scriptsize
\setlength{\tabcolsep}{4pt}
\renewcommand{\arraystretch}{1.12}
\begin{tabular}{
p{0.12\linewidth}
p{0.30\linewidth}
p{0.46\linewidth}
}
\toprule
\textbf{Style} & \textbf{Definition} & \textbf{Example} \\
\midrule
Brief
&
A short, compact instruction that keeps all necessary behavior while minimizing wording.
&
Remove private identifiers and unsafe links, normalize the remaining text, then keep only sanitized samples whose special-character share is below 15\%.
\\

Casual Chat
&
A casual chat-style request that sounds like a real user asking for help while still being complete enough to execute.
&
Hey, could you help me clean up these source files? I need the LaTeX comments and references removed, the macros expanded, and then please keep only sources with at least 300 words left.
\\

Data-First
&
The raw data appears before the concrete instruction, so the task request is weighted toward the end of the prompt.
&
Above is the raw document. Please remove links and contact information, normalize spacing, and then keep the document only if its final text has at least 500 characters.
\\

Exclusion-Based
&
A request centered on what must not remain in the output, useful for cleaning and filtering recipes.
&
When processing this report, make sure the final result contains no disclaimers, table residue, or lines longer than 180 characters. After that, judge whether it is suitable for retrieval.
\\

Goal-First
&
A prose description that emphasizes the intended final state rather than a numbered procedure.
&
The goal is to make these help documents clean enough for a support index, with links removed, repeated template sentences cleaned up, spacing made consistent, and only documents with no more than 20\% repeated words retained.
\\

Peer Handoff
&
A natural teammate-to-teammate handoff written in practical workplace language.
&
Could you take these LaTeX sources, remove the comments and bibliography, expand the simple macros, and then keep only sources whose final text length is at least 1,000 characters?
\\

Policy Rule
&
A formal processing requirement that reads like a data handling policy, not code.
&
Before release, the text must not contain emails, IP addresses, file paths, credentials, or long secret-like tokens; after sanitization, retain it only if the result still has at least 100 words.
\\

QC Filter
&
A request that sounds like data quality screening, emphasizing retention criteria and rejection conditions.
&
Quality-check this page after cleaning away links and noisy HTML\@. Keep it only if the remaining text has at least 50 words and looks usable as corpus content.
\\

Scenario Story
&
A scenario-style request that first explains the messy data situation and then asks for the needed processing.
&
I have raw crawl pages with markup, navigation links, and messy whitespace. Please turn them into clean readable text and keep only pages whose alphanumeric content makes up at least 60\% of the final text.
\\

Step-by-Step
&
A direct command that lists the required operations explicitly and in order.
&
Please clean this web page by removing the HTML, stripping links, normalizing the whitespace, and then keep it only if the cleaned text has at least 100 characters.
\\

Use-Case Framing
&
A task framed around a downstream use case such as retrieval, indexing, release, compliance, or corpus construction.
&
For downstream retrieval, process these reports so that disclaimers, table residue, and abnormal long lines are removed, then keep only reports whose longest remaining line is no more than 200 characters.
\\
\bottomrule
\end{tabular}
\caption{
Prompt styles for recipe verbalization. After a data refinement recipe and its deterministic Data-Juicer reference are fixed, we instantiate the same recipe under multiple user-facing styles. These styles vary the surface form and discourse context of the request while preserving the operator sequence, execution order, and filtering semantics.
}
\label{tab:prompt_styles}
\end{table*}

%% file: figs/judge_prompt.tex
\begin{figure*}[htbp]
\begin{tcolorbox}[
    colframe=black!80,
    colback=white,
    colbacktitle=black!80,
    coltitle=white,
    title={Judge Prompt for Instruction Validation},
    fonttitle=\bfseries,
    arc=1mm,
    boxrule=0.6pt,
    left=2mm, right=2mm, top=1mm, bottom=1mm
]
\begin{lstlisting}[style=promptstyle]
You are a strict benchmark prompt judge.
You will be given:
- The internal recipe definition and filter parameters.
- The operator code/doc evidence.
- One candidate user-facing prompt.
Judge whether the prompt is faithful to what the recipe actually does.
Mandatory keep conditions:
1. Functional equivalence: the prompt requests the same transformation and filtering behavior.
2. Order correctness: the requested order matches the internal recipe order.
3. No code leakage: the prompt does not mention operator names, parameter names, YAML, Python,
   hidden code, or implementation internals.
4. Threshold grounding: if the recipe has filter parameters, all active numeric thresholds must
   appear as natural user-facing constraints, not vague words like "long enough" or "too repetitive".
5. Wrapper compatibility: the user requirement can be safely combined with a fixed benchmark
   wrapper that separately supplies raw input text and the JSON output contract.
Also score:
- user_naturalness: does it sound like a plausible user request?
- threshold_grounding: are thresholds/conditions expressed naturally and correctly?
- clarity: is the request clear and executable?
- style_distinctiveness: does this candidate sound distinct from a generic template for the same style?
Return JSON only:
{
  "verdict": "keep" or "reject",
  "must_pass": {
    "functional_equivalence": true,
    "order_correct": true,
    "no_code_leakage": true,
    "thresholds_grounded": true,
    "wrapper_compatible": true
  },
  "scores": {
    "user_naturalness": 1-5,
    "threshold_grounding": 1-5,
    "clarity": 1-5,
    "style_distinctiveness": 1-5
  },
  "issues": ["..."],
  "summary": "one short sentence"
}
\end{lstlisting}
\end{tcolorbox}
\captionof{figure}{The LLM judge prompt used to validate candidate user instructions before they enter the final recipe-level prompt pool.}
\label{fig:judge_prompt}
\end{figure*}

%% file: figs/eval_prompt.tex
\begin{figure*}[htbp]
\begin{tcolorbox}[
    colframe=black!80,
    colback=white,
    colbacktitle=black!80,
    coltitle=white,
    title={Direct Execution Prompt (shared wrapper)},
    fonttitle=\bfseries,
    arc=1mm,
    boxrule=0.6pt,
    left=2mm, right=2mm, top=1mm, bottom=1mm
]
\begin{lstlisting}[style=promptstyle]
Task:
{user_requirement}
Raw input text: <input>{input_text}</input>
Return tagged output only.
Use exactly this format: {output_hint}
Rules:
- status must be KEEP or DROP inside <status>...</status>.
- Put the output text inside <clean_text>...</clean_text>.
- If status is KEEP, clean_text must be the final refined text.
- If status is DROP, clean_text must be the text state at the point where the sample is rejected.
- Preserve backslashes exactly as plain text; do not JSON-escape them.
- Do not output markdown, code fences, or explanations.
\end{lstlisting}
\end{tcolorbox}
\captionof{figure}{Direct Mode prompt, which also serves as the shared wrapper for Few-Shot, Plan-First, and State-Aware modes (Figures~\ref{fig:fewshot_prompt}--\ref{fig:state_aware_prompt}).}
\label{fig:direct_prompt}
\end{figure*}

\begin{figure*}[htbp]
\begin{tcolorbox}[
    colframe=black!80,
    colback=white,
    colbacktitle=black!80,
    coltitle=white,
    title={Few-Shot — examples prepended before the shared wrapper (Figure~\ref{fig:direct_prompt})},
    fonttitle=\bfseries,
    arc=1mm,
    boxrule=0.6pt,
    left=2mm, right=2mm, top=1mm, bottom=1mm
]
\begin{lstlisting}[style=promptstyle]
Example 1
Task:
{example_1_user_requirement}
Raw input text: <input>{example_1_input_text}</input>
Correct output:
<status>{example_1_reference_status}</status><clean_text>{example_1_reference_text}</clean_text>

Example 2
Task:
{example_2_user_requirement}
Raw input text: <input>{example_2_input_text}</input>
Correct output:
<status>{example_2_reference_status}</status><clean_text>{example_2_reference_text}</clean_text>
\end{lstlisting}
\end{tcolorbox}
\captionof{figure}{Few-Shot Mode: two solved examples from the same track are inserted in the shared wrapper. The shared wrapper (Figure~\ref{fig:direct_prompt}) follows unchanged.}
\label{fig:fewshot_prompt}
\end{figure*}

\begin{figure*}[htbp]
\begin{tcolorbox}[
    colframe=black!80,
    colback=white,
    colbacktitle=black!80,
    coltitle=white,
    title={Plan-First — analysis instruction appended to \texttt{\{user\_requirement\}}},
    fonttitle=\bfseries,
    arc=1mm,
    boxrule=0.6pt,
    left=2mm, right=2mm, top=1mm, bottom=1mm
]
\begin{lstlisting}[style=promptstyle]
Before cleaning the text, first write a short analysis block in <analyze>...</analyze>.
Use that block to restate the requested procedure as a concise standardized execution recipe.
List the steps in order. For each step, say what to do and include the relevant rule or
threshold when there is one.

Example analysis format:
<analyze>
Step 1: Remove repeated sentences.
Step 2: Normalize whitespace.
Step 3: Apply the word repetition filter with ratio <= 0.2, and drop the text if it fails.
</analyze>

Then execute the procedure on the text.
If a filter rejects the sample, stop at the rejection point and return DROP.
\end{lstlisting}
\end{tcolorbox}
\captionof{figure}{Plan-First Mode: this analysis instruction is appended to \texttt{\{user\_requirement\}} in the shared wrapper (Figure~\ref{fig:direct_prompt}), asking the model to restate the recipe as an ordered execution plan before producing output.}
\label{fig:plan_first_prompt}
\end{figure*}

\begin{figure*}[htbp]
\begin{tcolorbox}[
    colframe=black!80,
    colback=white,
    colbacktitle=black!80,
    coltitle=white,
    title={State-Aware — analysis instruction appended to \texttt{\{user\_requirement\}}},
    fonttitle=\bfseries,
    arc=1mm,
    boxrule=0.6pt,
    left=2mm, right=2mm, top=1mm, bottom=1mm
]
\begin{lstlisting}[style=promptstyle]
Before cleaning the text, first write a short analysis block in <analyze>...</analyze>.
Use that block to identify the intermediate text states that matter for correctness.
You may refer to them with names such as S0, S1, and S2.
Only text-changing steps should create a new state, while filter steps should be described
as operating on the relevant existing state.
Focus on which operation or filter should be applied to which state, especially when order
matters.

Example analysis format:
<analyze>
A useful state view is S0 = raw text and S1 = text after repeated-sentence removal.
The key risk is applying the repetition filter on S0 instead of S1; this filter should be
evaluated on S1 because the cleanup step changes the statistics.
</analyze>

In that analysis:
- refer only to the states or transitions most likely to change the result if used incorrectly;
- emphasize the specific state where an important filter or operation must be applied;
- do not analyze every step if most steps are unambiguous.

Then apply the intended procedure to the text.
If a filter rejects the sample, stop at the last valid state and return DROP.
\end{lstlisting}
\end{tcolorbox}
\captionof{figure}{State-Aware Mode: this analysis instruction is appended to \texttt{\{user\_requirement\}} in the shared wrapper (Figure~\ref{fig:direct_prompt}), asking the model to identify intermediate text states before executing the recipe.}
\label{fig:state_aware_prompt}
\end{figure*}

%% file: tables/real_scenario_appendix.tex
\begin{table}[h]
\centering
\resizebox{\columnwidth}{!}{
\begin{tabular}{@{}l ccccc ccc@{}}
\toprule
\textbf{Model} & Cont. & Loc. & Temp. & Ident. & Person & Avg & Comp & Gap \\
\midrule
Claude 4.7 Opus      & 90.1 & 70.6 & 83.8 & 63.2 & 71.9 & 75.9 & 50.6 & 25.3 \\
Claude 4.6 Sonnet    & 84.6 & 68.3 & 74.4 & 57.1 & 68.1 & 70.5 & 47.2 & 23.3 \\
Qwen3.6-Max          & 94.5 & 65.4 & 86.9 & 66.1 & 79.7 & 78.5 & 51.0 & 27.5 \\
GPT-5.5              & 89.7 & 67.6 & 76.9 & 66.1 & 57.7 & 71.6 & 51.8 & 19.8 \\
DeepSeek-V4-Flash    & 92.9 & 62.7 & 77.5 & 49.6 & 63.8 & 69.3 & 39.4 & 29.9 \\
Kimi-K2.6            & 91.3 & 65.0 & 83.8 & 59.3 & 59.5 & 71.8 & 34.2 & 37.6 \\
GLM-5                & 94.1 & 60.8 & 80.0 & 64.3 & 73.7 & 74.6 & 39.0 & 35.6 \\
Gemma-4              & 88.1 & 62.7 & 84.4 & 62.9 & 77.2 & 75.1 & 45.8 & 29.3 \\
Qwen3.6-27B          & 83.8 & 55.9 & 71.2 & 59.6 & 51.6 & 64.4 & 40.2 & 24.2 \\
Qwen3.6-35B-A3B      & 85.0 & 50.3 & 73.8 & 63.2 & 46.8 & 63.8 & 25.0 & 38.8 \\
\bottomrule
\end{tabular}
}
\caption{Full results on the PII semantic domain (RS@3 \%). Atomic subtasks: Contact (Cont.), Location (Loc.), Temporal (Temp.), Identification (Ident.), and Person. Avg is the macro-average across atomic subtasks. Comp is the compositional task. Gap is Avg $-$ Comp.}
\label{tab:semantic_pii_full}
\end{table}

\begin{table}[h]
\centering
\resizebox{\columnwidth}{!}{
\begin{tabular}{@{}l cccc ccc@{}}
\toprule
\textbf{Model} & Detect. & Span. & Type. & Correct. & Avg & Comp & Gap \\
\midrule
Claude 4.7 Opus      & 42.7 & 12.3 & 20.3 & 29.0 & 26.1 & 24.7 & 1.4 \\
Claude 4.6 Sonnet    & 41.3 & 17.3 & 14.3 & 29.7 & 25.7 & 24.7 & 1.0 \\
Qwen3.6-Max          & 43.0 & 22.7 & 33.0 & 35.3 & 33.5 & 27.7 & 5.8 \\
GPT-5.5              & 40.3 & 15.3 & 21.7 & 21.0 & 24.6 & 21.0 & 3.6 \\
DeepSeek-V4-Flash    & 53.7 & 33.3 & 32.0 & 48.0 & 41.8 & 36.3 & 5.4 \\
Kimi-K2.6            & 44.7 & 12.7 & 22.7 & 45.0 & 31.2 & 23.3 & 7.9 \\
GLM-5                & 47.7 & 32.0 & 31.7 & 37.0 & 37.1 & 29.0 & 8.1 \\
Gemma-4              & 43.7 & 26.0 & 37.3 & 29.7 & 34.2 & 29.3 & 4.8 \\
Qwen3.6-27B          & 51.0 & 26.7 & 40.3 & 32.7 & 37.7 & 35.0 & 2.7 \\
Qwen3.6-35B-A3B      & 51.7 & 20.0 & 33.7 & 38.3 & 35.9 & 37.0 & $-$1.1 \\
\bottomrule
\end{tabular}
}
\caption{Full results on the hallucination semantic domain (RS@3 \%). Atomic subtasks: Detection (Detect.), Span extraction (Span.), Type classification (Type.), and Correction (Correct.). Avg is the macro-average across atomic subtasks. Comp uses the mixed structured-text schema requiring detection, spans, types, and corrected text in one output. Gap is Avg $-$ Comp.}
\label{tab:semantic_hallu_full}
\end{table}

\begin{table}[h]
\centering
\resizebox{\columnwidth}{!}{
\begin{tabular}{@{}l ccc ccc@{}}
\toprule
\textbf{Model} & Prom. & Resp. & Cate. & Avg & Comp & Gap \\
\midrule
Claude 4.7 Opus      & 84.0 & 89.0 & 58.0 & 77.0 & 48.0 & 29.0 \\
Claude 4.6 Sonnet    & 83.7 & 87.3 & 54.0 & 75.0 & 46.0 & 29.0 \\
Qwen3.6-Max          & 88.0 & 86.3 & 57.7 & 77.3 & 45.0 & 32.3 \\
GPT-5.5              & 86.7 & 74.3 & 55.0 & 72.0 & 46.0 & 26.0 \\
DeepSeek-V4-Flash    & 89.7 & 82.3 & 56.3 & 76.1 & 49.0 & 27.1 \\
Kimi-K2.6            & 80.0 & 85.3 & 57.3 & 74.2 & 49.3 & 24.9 \\
GLM-5                & 86.0 & 85.3 & 55.3 & 75.6 & 50.3 & 25.2 \\
Gemma-4              & 83.0 & 83.7 & 50.7 & 72.4 & 44.7 & 27.8 \\
Qwen3.6-27B          & 88.0 & 91.7 & 58.3 & 79.3 & 52.7 & 26.7 \\
Qwen3.6-35B-A3B      & 88.3 & 87.3 & 54.7 & 76.8 & 48.3 & 28.4 \\
\bottomrule
\end{tabular}
}
\caption{Full results on the safety-tagging semantic domain (RS@3 \%). Atomic subtasks: prompt-label classification (Prom.), response-label classification (Resp.), and violated-category tagging (Cate.). Avg is the macro-average across atomic subtasks. Comp requires all three fields in one JSON object. Gap is Avg $-$ Comp.}
\label{tab:semantic_safety_full}
\end{table}

\begin{table}[h]
\centering
\resizebox{\columnwidth}{!}{
\begin{tabular}{@{}l ccccc ccc@{}}
\toprule
\textbf{Model} & Help. & Correct. & Coher. & Complex. & Verb. & Avg & Comp & Gap \\
\midrule
Claude 4.7 Opus      & 33.0 & 36.7 & 64.0 & 62.0 & 60.7 & 51.3 & 7.7 & 43.6 \\
Claude 4.6 Sonnet    & 35.0 & 34.0 & 49.3 & 52.7 & 42.7 & 42.7 & 2.7 & 40.1 \\
Qwen3.6-Max          & 53.7 & 51.0 & 71.0 & 60.0 & 72.0 & 61.5 & 6.3 & 55.2 \\
GPT-5.5              & 39.3 & 44.0 & 68.7 & 55.0 & 44.0 & 50.2 & 5.7 & 44.5 \\
DeepSeek-V4-Flash    & 57.0 & 58.3 & 78.7 & 69.3 & 60.3 & 64.7 & 4.0 & 60.7 \\
Kimi-K2.6            & 38.3 & 51.3 & 59.3 & 72.3 & 38.0 & 51.9 & 1.3 & 50.5 \\
GLM-5                & 48.7 & 54.0 & 75.7 & 38.0 & 27.3 & 48.7 & 2.3 & 46.4 \\
Gemma-4              & 49.0 & 48.7 & 72.3 & 38.7 & 68.0 & 55.3 & 5.3 & 50.0 \\
Qwen3.6-27B          & 53.0 & 52.0 & 74.3 & 71.7 & 61.0 & 62.4 & 3.7 & 58.7 \\
Qwen3.6-35B-A3B      & 53.7 & 56.7 & 66.0 & 76.0 & 63.3 & 63.1 & 2.0 & 61.1 \\
\bottomrule
\end{tabular}
}
\caption{Full results on the rubric-scoring semantic domain (RS@3 \%). Atomic subtasks correspond to HelpSteer2 dimensions: Helpfulness (Help.), Correctness (Correct.), Coherence (Coher.), Complexity (Complex.), and Verbosity (Verb.). Avg is the macro-average across atomic subtasks. Comp requires all five scores in one JSON object. Gap is Avg $-$ Comp.}
\label{tab:semantic_rubric_full}
\end{table}

%% file: figs/semantic_prompts.tex
% ── Semantic Track Prompt Templates ──

\begin{figure*}[ht]
\begin{tcolorbox}[
    % breakable,
    colframe=black!80,
    colback=white,
    colbacktitle=black!80,
    coltitle=white,
    title={System Prompt (shared)},
    fonttitle=\bfseries,
    arc=1mm,
    boxrule=0.6pt,
    left=2mm, right=2mm, top=1mm, bottom=1mm
]
\begin{lstlisting}[style=promptstyle]
You are a careful data refinement engine. Follow the user request exactly.
Return only the required output. Do not explain your reasoning.
\end{lstlisting}
\end{tcolorbox}

\begin{tcolorbox}[
    % breakable,
    colframe=black!80,
    colback=white,
    colbacktitle=black!80,
    coltitle=white,
    title={User Prompt Wrapper (text tasks: PII redaction, Hallu correction)},
    fonttitle=\bfseries,
    arc=1mm,
    boxrule=0.6pt,
    left=2mm, right=2mm, top=1mm, bottom=1mm
]
\begin{lstlisting}[style=promptstyle]
Task:
{user_requirement}

Raw input text:
<input>
{input_text}
</input>

Return tagged output only.
Use exactly this format: <status>KEEP</status><clean_text>...</clean_text>
Rules:
- status must be KEEP.
- Put the processed text inside <clean_text>...</clean_text>.
- Do not output markdown, code fences, or explanations.
\end{lstlisting}
\end{tcolorbox}

\begin{tcolorbox}[
    % breakable,
    colframe=black!80,
    colback=white,
    colbacktitle=black!80,
    coltitle=white,
    title={User Prompt Wrapper (structured JSON tasks: hallucination diagnosis, rubric scoring, safety tagging)},
    fonttitle=\bfseries,
    arc=1mm,
    boxrule=0.6pt,
    left=2mm, right=2mm, top=1mm, bottom=1mm
]
\begin{lstlisting}[style=promptstyle]
Task:
{user_requirement}

Input text:
<input>
{input_text}
</input>

Return only the JSON output. Do not wrap in code fences or add explanations.
\end{lstlisting}
\end{tcolorbox}
\captionof{figure}{Shared system prompt and user prompt wrappers for all semantic-extension domain evaluations.}
\label{fig:semantic_prompt_wrappers}
\end{figure*}

\begin{figure*}[ht]
\begin{tcolorbox}[
    % breakable,
    colframe=black!80,
    colback=white,
    colbacktitle=black!80,
    coltitle=white,
    title={PII Atomic — \texttt{user\_requirement} (example: contact group, direct style)},
    fonttitle=\bfseries,
    arc=1mm,
    boxrule=0.6pt,
    left=2mm, right=2mm, top=1mm, bottom=1mm
]
\begin{lstlisting}[style=promptstyle]
You are a data refinement engine. Apply the following PII redaction steps to
the input text in order.

Step 1: Identify and replace all contact information (email addresses,
telephone numbers) with placeholders. Use the format [LABEL_N] where LABEL is
the original entity type (one of: EMAIL, TELEPHONENUM) and N is a counter
starting from 1 for each label type. Leave all other text unchanged.

Return the result in the format: <status>KEEP</status><clean_text>...</clean_text>
\end{lstlisting}
\end{tcolorbox}

\begin{tcolorbox}[
    % breakable,
    colframe=black!80,
    colback=white,
    colbacktitle=black!80,
    coltitle=white,
    title={PII Compositional — \texttt{user\_requirement} (example: 4 groups, direct style)},
    fonttitle=\bfseries,
    arc=1mm,
    boxrule=0.6pt,
    left=2mm, right=2mm, top=1mm, bottom=1mm
]
\begin{lstlisting}[style=promptstyle]
You are a data refinement engine. Apply the following PII redaction steps to
the input text in order.

Step 1: Identify and replace all contact information (email addresses,
telephone numbers) with placeholders. Use the format [LABEL_N] where LABEL is
the original entity type (one of: EMAIL, TELEPHONENUM) and N is a counter
starting from 1 for each label type. Leave all other text unchanged.
Step 2: Identify and replace all identification numbers (ID cards, bank
accounts, social security, tax IDs, driver licenses, credit cards) with
placeholders. Use the format [LABEL_N] where LABEL is the original entity
type (one of: IDCARDNUM, ACCOUNTNUM, SOCIALNUM, TAXNUM, DRIVERLICENSENUM,
CREDITCARDNUMBER) and N is a counter starting from 1 for each label type.
Leave all other text unchanged.
Step 3: Identify and replace all location information (cities, street
addresses, building numbers, zip codes) with placeholders. ...
Step 4: Identify and replace all person names and usernames (given names,
surnames, usernames) with placeholders. ...

Return the result in the format: <status>KEEP</status><clean_text>...</clean_text>
\end{lstlisting}
\end{tcolorbox}
\captionof{figure}{PII redaction prompt templates. Atomic shows a single-group instruction; compositional enumerates all present groups as ordered steps. The \texttt{direct} style is shown.}
\label{fig:semantic_pii_prompts}
\end{figure*}

\begin{figure*}[ht]
\begin{tcolorbox}[
    % breakable,
    colframe=black!80,
    colback=white,
    colbacktitle=black!80,
    coltitle=white,
    title={Hallu Atomic: Detection — \texttt{user\_requirement} (direct style)},
    fonttitle=\bfseries,
    arc=1mm,
    boxrule=0.6pt,
    left=2mm, right=2mm, top=1mm, bottom=1mm
]
\begin{lstlisting}[style=promptstyle]
Inspect the given text and determine whether it contains any hallucinated,
fabricated, or factually unsupported content. Return a JSON object with
exactly these keys: {"has_hallucination": true/false, "hallucination_count": N}.
\end{lstlisting}
\end{tcolorbox}

\begin{tcolorbox}[
    % breakable,
    colframe=black!80,
    colback=white,
    colbacktitle=black!80,
    coltitle=white,
    title={Hallu Atomic: Span Extraction — \texttt{user\_requirement} (direct style)},
    fonttitle=\bfseries,
    arc=1mm,
    boxrule=0.6pt,
    left=2mm, right=2mm, top=1mm, bottom=1mm
]
\begin{lstlisting}[style=promptstyle]
Identify all hallucinated text spans in the given text. For each span, extract
the exact text that is hallucinated and its hallucination type. Return a JSON
object: {"spans": [{"text": "...", "type": "..."}]}.
\end{lstlisting}
\end{tcolorbox}

\begin{tcolorbox}[
    % breakable,
    colframe=black!80,
    colback=white,
    colbacktitle=black!80,
    coltitle=white,
    title={Hallu Atomic: Type Classification — \texttt{user\_requirement} (direct style)},
    fonttitle=\bfseries,
    arc=1mm,
    boxrule=0.6pt,
    left=2mm, right=2mm, top=1mm, bottom=1mm
]
\begin{lstlisting}[style=promptstyle]
Classify the hallucination types present in the given text. Possible types:
entity, relation, subjective, contradictory, invented, unverifiable,
fictional, format, other.
Return a JSON object: {"types": ["type1", "type2", ...]}.
\end{lstlisting}
\end{tcolorbox}

% \begin{tcolorbox}[
%     breakable,
%     colframe=black!80,
%     colback=white,
%     colbacktitle=black!80,
%     coltitle=white,
%     title={Hallu Atomic: Correction — \texttt{user\_requirement} (direct style)},
%     fonttitle=\bfseries,
%     arc=1mm,
%     boxrule=0.6pt,
%     left=2mm, right=2mm, top=1mm, bottom=1mm
% ]
% \begin{lstlisting}[style=promptstyle]
% Correct all hallucinated content in the given text. If the text contains NO hallucinations, return it EXACTLY as is without any changes do not rephrase,
% reword, or reformat it. If hallucinations are present, remove fabricated,
% contradictory, or unverifiable statements and replace incorrect information
% with accurate corrections where available. Preserve all factual content
% unchanged.
% Return: <status>KEEP</status><clean_text>corrected text</clean_text>
% \end{lstlisting}
% \end{tcolorbox}

\begin{tcolorbox}[
    % breakable,
    colframe=black!80,
    colback=white,
    colbacktitle=black!80,
    coltitle=white,
    title={Hallu Atomic: Correction — \texttt{user\_requirement} (direct style)},
    fonttitle=\bfseries,
    arc=1mm,
    boxrule=0.6pt,
    left=2mm, right=2mm, top=1mm, bottom=1mm
]
\begin{lstlisting}[style=promptstyle]
Correct all hallucinated content in the given text. If the text contains NO
hallucinations, copy the input text exactly as corrected_text without
rephrasing, rewording, or reformatting it. If hallucinations are present,
remove fabricated, contradictory, or unverifiable statements and replace
incorrect information with accurate corrections where available. Preserve all
factual content unchanged.

Return exactly a JSON object: {"corrected_text": "..."}
\end{lstlisting}
\end{tcolorbox}

% \begin{tcolorbox}[
%     breakable,
%     colframe=black!80,
%     colback=white,
%     colbacktitle=black!80,
%     coltitle=white,
%     title={Hallu Compositional — \texttt{user\_requirement} (direct style)},
%     fonttitle=\bfseries,
%     arc=1mm,
%     boxrule=0.6pt,
%     left=2mm, right=2mm, top=1mm, bottom=1mm
% ]
% \begin{lstlisting}[style=promptstyle]
% You are a hallucination detection and correction engine. Analyze the given
% text and perform the following steps:
% 1. Detect whether the text contains hallucinated content.
% 2. If NO hallucinations are found, return the text EXACTLY as-is without any
%    changes.
% 3. If hallucinations exist, identify all hallucinated spans and their types
%    (entity/relation/subjective/contradictory/invented/unverifiable).
% 4. Correct all hallucinated content by removing fabrications and applying
%    known corrections. Preserve all factual content unchanged.
% Return the corrected text: <status>KEEP</status><clean_text>corrected text</clean_text>
% \end{lstlisting}
% \end{tcolorbox}

\begin{tcolorbox}[
    % breakable,
    colframe=black!80,
    colback=white,
    colbacktitle=black!80,
    coltitle=white,
    title={Hallu Compositional — \texttt{user\_requirement} (direct style)},
    fonttitle=\bfseries,
    arc=1mm,
    boxrule=0.6pt,
    left=2mm, right=2mm, top=1mm, bottom=1mm
]
\begin{lstlisting}[style=promptstyle]
You are a hallucination analysis and correction engine. Analyze the given text
and return one JSON object with exactly these keys:
{"has_hallucination": true/false, "hallucination_count": N,
 "spans": [{"text": "...", "type": "..."}],
 "types": ["..."], "corrected_text": "..."}.

First detect whether hallucinations are present, then extract every
hallucinated span with its type, then list the unique hallucination types,
then produce corrected_text. If no hallucinations are present, use false, 0,
empty arrays, and copy the input text exactly as corrected_text.
\end{lstlisting}
\end{tcolorbox}
\captionof{figure}{Hallucination prompt templates. Atomic decomposes into four independently evaluated sub-tasks: detection, span extraction, type classification, and correction. Compositional requires the full pipeline in a single JSON output containing detection fields, hallucinated spans, hallucination types, and corrected text. The \texttt{direct} style is shown.}
\label{fig:semantic_hallu_prompts}
\end{figure*}

\begin{figure*}[ht]
\begin{tcolorbox}[
    % breakable,
    colframe=black!80,
    colback=white,
    colbacktitle=black!80,
    coltitle=white,
    title={Safety Atomic: Prompt Label — \texttt{user\_requirement} (direct style)},
    fonttitle=\bfseries,
    arc=1mm,
    boxrule=0.6pt,
    left=2mm, right=2mm, top=1mm, bottom=1mm
]
\begin{lstlisting}[style=promptstyle]
Classify the requested Aegis safety field: prompt_label. Labels: safe or
unsafe. Use unsafe when the prompt or response violates one or more Aegis
categories. Categories: Needs Caution, Hate/Identity Hate, Sexual, Suicide
and Self Harm, Violence, Guns/Illegal Weapons, Threat, PII/Privacy, Sexual
Minor, Criminal Planning/Confessions, Harassment, Controlled/Regulated
Substances, Profanity. Return exactly this JSON shape:
{"prompt_label": "safe or unsafe"}.
\end{lstlisting}
\end{tcolorbox}

\begin{tcolorbox}[
    % breakable,
    colframe=black!80,
    colback=white,
    colbacktitle=black!80,
    coltitle=white,
    title={Safety Atomic: Response Label — \texttt{user\_requirement} (direct style)},
    fonttitle=\bfseries,
    arc=1mm,
    boxrule=0.6pt,
    left=2mm, right=2mm, top=1mm, bottom=1mm
]
\begin{lstlisting}[style=promptstyle]
Classify the requested Aegis safety field: response_label. Labels: safe or
unsafe. Use unsafe when the prompt or response violates one or more Aegis
categories. Categories: Needs Caution, Hate/Identity Hate, Sexual, Suicide
and Self Harm, Violence, Guns/Illegal Weapons, Threat, PII/Privacy, Sexual
Minor, Criminal Planning/Confessions, Harassment, Controlled/Regulated
Substances, Profanity. Return exactly this JSON shape:
{"response_label": "safe or unsafe"}.
\end{lstlisting}
\end{tcolorbox}

\begin{tcolorbox}[
    % breakable,
    colframe=black!80,
    colback=white,
    colbacktitle=black!80,
    coltitle=white,
    title={Safety Atomic: Violated Categories — \texttt{user\_requirement} (direct style)},
    fonttitle=\bfseries,
    arc=1mm,
    boxrule=0.6pt,
    left=2mm, right=2mm, top=1mm, bottom=1mm
]
\begin{lstlisting}[style=promptstyle]
Classify the requested Aegis safety field: violated_categories. Labels: safe
or unsafe. Use unsafe when the prompt or response violates one or more Aegis
categories. Categories: Needs Caution, Hate/Identity Hate, Sexual, Suicide
and Self Harm, Violence, Guns/Illegal Weapons, Threat, PII/Privacy, Sexual
Minor, Criminal Planning/Confessions, Harassment, Controlled/Regulated
Substances, Profanity. Return exactly this JSON shape:
{"violated_categories": ["category names"]}.
\end{lstlisting}
\end{tcolorbox}

\begin{tcolorbox}[
    % breakable,
    colframe=black!80,
    colback=white,
    colbacktitle=black!80,
    coltitle=white,
    title={Safety Compositional — \texttt{user\_requirement} (direct style)},
    fonttitle=\bfseries,
    arc=1mm,
    boxrule=0.6pt,
    left=2mm, right=2mm, top=1mm, bottom=1mm
]
\begin{lstlisting}[style=promptstyle]
Classify the Aegis safety labels for the user prompt and assistant response.
Labels: safe or unsafe. Use unsafe when the prompt or response violates one
or more Aegis categories. Categories: Needs Caution, Hate/Identity Hate,
Sexual, Suicide and Self Harm, Violence, Guns/Illegal Weapons, Threat,
PII/Privacy, Sexual Minor, Criminal Planning/Confessions, Harassment,
Controlled/Regulated Substances, Profanity. Return exactly this JSON shape:
{"prompt_label": "safe or unsafe", "response_label": "safe or unsafe",
 "violated_categories": ["category names"]}.
\end{lstlisting}
\end{tcolorbox}

\captionof{figure}{Safety tagging prompt templates. Atomic tasks classify one Aegis field at a time, while the compositional task requires prompt label, response label, and violated categories in one JSON object. The \texttt{direct} style is shown.}
\label{fig:semantic_safety_prompts}
\end{figure*}

\begin{figure*}[ht]
\begin{tcolorbox}[
    % breakable,
    colframe=black!80,
    colback=white,
    colbacktitle=black!80,
    coltitle=white,
    title={Rubric Atomic: Helpfulness — \texttt{user\_requirement} (direct style)},
    fonttitle=\bfseries,
    arc=1mm,
    boxrule=0.6pt,
    left=2mm, right=2mm, top=1mm, bottom=1mm
]
\begin{lstlisting}[style=promptstyle]
Score the assistant response for helpfulness. In HelpSteer2, helpfulness
means overall helpfulness of the response to the prompt. Use the 0-4 rubric
scale. Return exactly this JSON shape: {"helpfulness": "integer 0-4"}.
\end{lstlisting}
\end{tcolorbox}

\begin{tcolorbox}[
    % breakable,
    colframe=black!80,
    colback=white,
    colbacktitle=black!80,
    coltitle=white,
    title={Rubric Atomic: Correctness — \texttt{user\_requirement} (direct style)},
    fonttitle=\bfseries,
    arc=1mm,
    boxrule=0.6pt,
    left=2mm, right=2mm, top=1mm, bottom=1mm
]
\begin{lstlisting}[style=promptstyle]
Score the assistant response for correctness. In HelpSteer2, correctness
means whether the response includes pertinent facts without errors. Use the
0-4 rubric scale. Return exactly this JSON shape:
{"correctness": "integer 0-4"}.
\end{lstlisting}
\end{tcolorbox}

\begin{tcolorbox}[
    % breakable,
    colframe=black!80,
    colback=white,
    colbacktitle=black!80,
    coltitle=white,
    title={Rubric Atomic: Coherence — \texttt{user\_requirement} (direct style)},
    fonttitle=\bfseries,
    arc=1mm,
    boxrule=0.6pt,
    left=2mm, right=2mm, top=1mm, bottom=1mm
]
\begin{lstlisting}[style=promptstyle]
Score the assistant response for coherence. In HelpSteer2, coherence means
consistency and clarity of expression. Use the 0-4 rubric scale. Return
exactly this JSON shape: {"coherence": "integer 0-4"}.
\end{lstlisting}
\end{tcolorbox}

\begin{tcolorbox}[
    % breakable,
    colframe=black!80,
    colback=white,
    colbacktitle=black!80,
    coltitle=white,
    title={Rubric Atomic: Complexity — \texttt{user\_requirement} (direct style)},
    fonttitle=\bfseries,
    arc=1mm,
    boxrule=0.6pt,
    left=2mm, right=2mm, top=1mm, bottom=1mm
]
\begin{lstlisting}[style=promptstyle]
Score the assistant response for complexity. In HelpSteer2, complexity means
intellectual depth required to write the response. Use the 0-4 rubric scale.
Return exactly this JSON shape: {"complexity": "integer 0-4"}.
\end{lstlisting}
\end{tcolorbox}

\begin{tcolorbox}[
    % breakable,
    colframe=black!80,
    colback=white,
    colbacktitle=black!80,
    coltitle=white,
    title={Rubric Atomic: Verbosity — \texttt{user\_requirement} (direct style)},
    fonttitle=\bfseries,
    arc=1mm,
    boxrule=0.6pt,
    left=2mm, right=2mm, top=1mm, bottom=1mm
]
\begin{lstlisting}[style=promptstyle]
Score the assistant response for verbosity. In HelpSteer2, verbosity means
amount of detail included relative to what the prompt asks for. Use the 0-4
rubric scale. Return exactly this JSON shape:
{"verbosity": "integer 0-4"}.
\end{lstlisting}
\end{tcolorbox}

\begin{tcolorbox}[
    % breakable,
    colframe=black!80,
    colback=white,
    colbacktitle=black!80,
    coltitle=white,
    title={Rubric Compositional — \texttt{user\_requirement} (direct style)},
    fonttitle=\bfseries,
    arc=1mm,
    boxrule=0.6pt,
    left=2mm, right=2mm, top=1mm, bottom=1mm
]
\begin{lstlisting}[style=promptstyle]
Score the assistant response on all HelpSteer2 rubric dimensions.
Definitions: helpfulness: overall helpfulness of the response to the prompt;
correctness: whether the response includes pertinent facts without errors;
coherence: consistency and clarity of expression; complexity: intellectual
depth required to write the response; verbosity: amount of detail included
relative to what the prompt asks for. Return exactly this JSON shape:
{"helpfulness": "integer 0-4", "correctness": "integer 0-4",
 "coherence": "integer 0-4", "complexity": "integer 0-4",
 "verbosity": "integer 0-4"}.
\end{lstlisting}
\end{tcolorbox}

\captionof{figure}{Rubric scoring prompt templates. Atomic tasks score one HelpSteer2 dimension at a time, while the compositional task requires all five rubric scores in one JSON object. The \texttt{direct} style is shown.}
\label{fig:semantic_rubric_prompts}
\end{figure*}